# Nonparametric Latent Tree Graphical Models: Inference, Estimation, and Structure Learning

Le Song[*], Han Liu[†], Ankur Parikh[‡], and Eric Xing[§]

**Abstract**

Tree structured graphical models are powerful at expressing long range or hierarchical dependency among many variables, and have been widely applied in different areas of computer science and statistics. However, existing methods for parameter estimation, inference, and structure learning mainly rely on the Gaussian or discrete assumptions, which are restrictive under many applications. In this paper, we propose new nonparametric methods based on reproducing kernel Hilbert space embeddings of distributions that can recover the latent tree structures, estimate the parameters, and perform inference for high dimensional continuous and non-Gaussian variables. The usefulness of the proposed methods are illustrated by thorough numerical results.

## 1  Introduction

Modern data acquisition routinely produces massive amounts of high dimensional data with complex statistical dependency structures. Latent variable graphical models provide a succinct representation of such complex dependency structures by relating the observed variables to a set of latent ones. By defining a joint distribution over observed and latent variables, the marginal distribution of the observed variables can be obtained by integrating out the latent ones. This allows complex distributions over observed variables (e.g., clique models) to be expressed in terms of more tractable joint models (e.g., tree models) over the augmented variable space. Probabilistic graphical models with latent variables have been deployed successfully to a diverse range of problems such as in document analysis (Blei et al., 2002), social network modeling (Hoff et al., 2002), speech recognition (Rabiner and Juang, 1986) and bioinformatics (Clark, 1990).

In this paper, we focus on latent variable models where the latent structures are trees (we call it a "latent tree" for short). In these tree-structured graphical models, the leaves are the set of observed variables (e.g., taxa, pixels, words) while the internal nodes are hidden and intuitively "represent" the common properties of their descendants (e.g., distinct ancestral species, objects in an image, latent semantics). This class of models strike a nice balance between their representation power (e.g., it can still model cliques) and the complexity of learning and inference processes on these structures (e.g., the message passing inference algorithm is exact on trees). In particular, we propose a unified nonparametric framework for addressing three key questions in tree-structured latent variable models:

- Estimating latent tree structures

- Estimate model parameters

- Conduct inference on the obtained tree graphical models.

[*]College of Computing, Georgia Institute of Technology, Atlanta, Georgia, 30332, USA; email: lsong@cc.gatech.edu

[†]Department of Operations Research and Financial Engineering, Princeton University, Princeton, NJ 08544, USA; e-mail: hanliu@princeton.edu

[‡]School of Computer Science, Carnegie Mellon University, Pittsburgh, PA, 15213, USA; email: aparikh@cs.cmu.edu

[§]School of Computer Science, Carnegie Mellon University, Pittsburgh, PA, 15213, USA; email: epxing@cs.cmu.edu

In the current literature, the problem of estimating the structure of latent trees has largely been tackled by heuristic algorithms since exhaustively searching over the whole latent tree space is intractable. For instance, Zhang (2004) proposed a search heuristic for hierarchical latent class models by defining a series of local search operations and using the expectation-maximization (EM) algorithm to compute the likelihood of candidate structures. Harmeling and Williams (2010) proposed a greedy algorithm to learn binary trees by joining two nodes with a high mutual information and iteratively performing the EM algorithm to compute the mutual information among newly added hidden nodes. Alternatively, Heller and Ghahramani (2005) proposed the Bayesian hierarchical clustering method, which is an agglomerative clustering technique that merges clusters based on statistical hypothesis testing. Many other local search heuristics based on maximum parsimony and maximum likelihood methods can also be found from the phylogenetic community (Semple and Steel, 2003). However, none of these methods extends easily to the nonparametric setting since they require the data to be either discrete or at least the distributions have a parametric form such that likelihood based testing and estimation can be easily computed.

Besides these heuristic approaches, many methods with provable guarantees have also been proposed in the phylogenetic community (Erdös et al., 1999a,b; Mossel and Roch, 2006; Mossel, 2007; Roch, 2010; Mossel et al., 2011a,b; Gronau et al., 2008; Daskalakis et al., 2006, 2011; Anandkumar et al., 2011; Anandkumar and Valluvan, 2013). For instnace, neighbor joining algorithm (Saitou et al., 1987) and recursive grouping algorithm (Choi et al., 2010) take a pairwise distance matrix between all observed variables as input and output an estimated tree graph by iteratively adding hidden nodes. While these methods are iterative, they have strong theoretical guarantees on structure recovery when the true distance matrix forms an additive tree metric, i.e., the distance between nodes $s$ and $t$ is equal to the sum of the distances of the edges along its path in a tree $T$. In this case these methods are guaranteed to recover the correct structure (binary for neighbor joining, but arbitrary branching for the recursive grouping method). However, a major challenge in applying these distance based methods to the continuous, nonparametric case is that it is not clear how to define a valid additive tree metric, and after recovering the latent tree structure, how to perform efficient parameter learning and probabilistic inference (e.g., calculating the marginal or conditional distributions).

Once the latent tree structure is correctly recovered, estimating the model parameters has predominantly relied on likelihood maximization and local search heuristics such as the EM algorithm (Dempster et al., 1977). Besides the problem of local minima, nonGaussian statistical features such as multimodality and skewness may pose additional challenge for EM. For instance, parametric models such as mixture of Gaussians may lead to an exponential blowup in terms of their representation during the inference stage of the EM algorithm. Further approximations are needed to make the computation tractable. These approximations make it even harder to analyze the theoretical properties of the obtained estimator. In addition, the EM algorithm in general requires many iterations to reach a prespecified training precision.

In this paper, we all propose a method for learning the parameters of tree-structured latent variable models with continuous and non-Gaussian observation based on the concept of Hilbert space embedding of distributions (Smola et al., 2007a). The problems we try to address include: (1) How to estimate the structures of latent trees with strong theoretical guarantees; (2) How to perform local-minimum-free learning and inference based on the estimated tree structures, all in nonparametric setting. The main idea of our method is to exploit the spectral properties of the joint embedding (or covariance operators) in all the structure recovery, parameter learning, and probabilistic inference stages. For structure learning, we define a distance measure between pairwise variables based on singular value decomposition of covariance operators. This allows us to generalize existing distance based latent tree learning procedures such as neighbor joining (Saitou et al., 1987) and recursive grouping (Choi et al., 2010) to fully nonparametric settings. In this paper, we focus on the case where the set of observed variables (leaves) are continuous-valued and their joint distributions can not be easily characterized by a parametric family (e.g., Gaussian). Thus our approach fundamentally differs from almost all others, which mainly consider the case of discrete or Gaussian variables. Similar to their parametric counterparts, the obtained structure learning algorithms have strong theoretical guarantees. After tree structures have been recovered, we further exploit the principal singular vectors of the covariance operator to estimate the parameters of the latent variables. One advantage of our spectral algorithm is that it is local-minimum-free and hence amenable for further theoretical analysis. In particular, we will demonstrate the advantage of our method over existing approaches in both simulation and real data experiments.

The rest of this paper is organized as follows. First, we will explain our terminology of latent tree graphical models in Section 2. In Section 3, we will provide a roadmap of our nonparametric framework which uses kernel embedding of distributions as key technique. In Section 4, we will present connection between kernel density estimation for tree-structured latent variable models and kernel embedding of the distributions of such models. Then in Section 5, we will first explain our nonparametric methods for inference and parameter learning. And in Section 6, a nonparametric method for structure learning, both based on the idea of Hilbert space embedding of distributions. Last, we will present our experimental results in synthetic and real datasets in Section 7. An introduction of the concept of Hilbert space embedding of distributions and conditional distributions is provided in Appendix Section A and B respectively. Furthermore, the connection between Hilbert space embedding of distributions and higher order tensors is presented in Section C instead.

## 2  Latent Tree Graphical Models

In this paper, we focus on latent variable models where the observed variables are continuous nonGaussian. We also assume the conditional independence structures of the joint distribution of the observed and latent variables are fully specified by trees. We will use uppercase letters to denote random variables (e.g., $X_i$) and lowercase letters to denote the corresponding realizations (e.g., $x_i$). For notational simplicity, we assume that the domain $\Omega$ of all variables are the same. Generalization to the cases where the variables have different domains is straightforward. A latent tree model defines a joint distribution over a set of $O$ observed variables, denoted by $\mathscr{O} = \{X_1, \ldots, X_O\}$, and a set of $H$ hidden variables, denoted by $\mathscr{H} = \{X_{O+1}, \ldots, X_{O+H}\}$. The complete set of variables is denoted by $\mathscr{X} = \mathscr{O} \cup \mathscr{H}$. For simplicity, we assume that

(A1) All observed variables have the same domain $\mathcal{X}_\mathscr{O}$, and all hidden variables take from at most $k$ values from $\mathcal{X}_\mathscr{H}$. Furthermore, all observed variables are leaf nodes of the tree, and each hidden node in the tree has exactly 3 neighbors.

In this case, after we reroot the tree and redirect all the edges, for a node $s$ (corresponding to the variable $X_s$), we use $\alpha_s$ to denote its sibling, $\pi_s$ to denote its parent, $\iota_s$ to denote its left child and $\rho_s$ to denote its right child; the root node will have 3 children, we use $\omega_s$ to denote the extra child of the root node. All the observed variables are leaves in the tree, and we will use $\iota_s^*$, $\rho_s^*$, $\pi_s^*$ to denote an arbitrary observed variable which is found by tracing in the direction from node $s$ to its left child $\iota_s$, right child $\rho_s$, and its parent $\pi_s$ respectively. See Figure 1 for a summary of the notation.

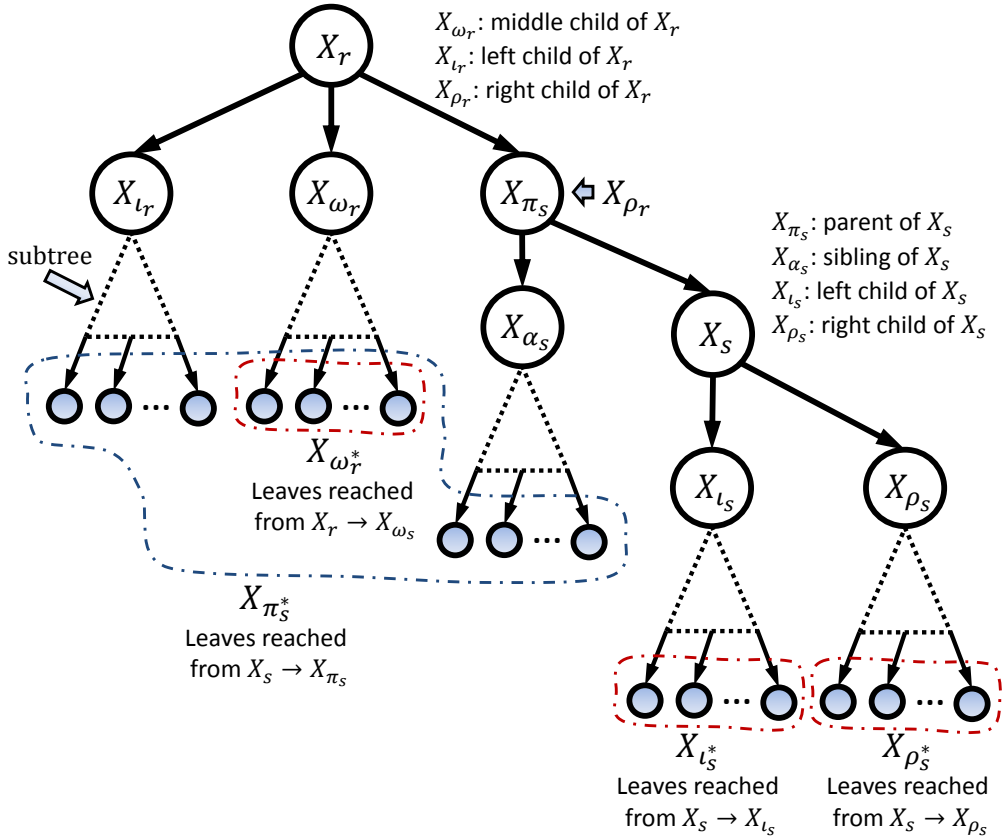

Figure 1: Illustration of a tree-structured latent variable model (or latent tree model). Open circle denotes hidden variables, and filled circles for observaed ones. For a node $s$, we use $\alpha_s$ to denote its sibling, $\pi_s$ to denote its parent, $\iota_s$ to denote its left child and $\rho_s$ to denote its right child; the root node will have 3 children, we use $\omega_s$ to denote the extra child.

The joint distribution of $\mathscr{X}$ in a latent tree model is fully characterized by a set of conditional distributions. More specifically, we can select an arbitrary latent node in the tree as the root, and reorient all edges away from the root. Then the set of conditional distributions between nodes and their parents $p(X_i|X_{\pi_i})$ are sufficient to characterize the joint distribution (for the root node $X_r$, we set $p(X_r|X_{\pi_r}) = p(X_r)$) and use $p(\cdot)$ to refer to density in continuous case with

$$p(\mathscr{X}) = \prod_{i=1}^{O+H} p(X_i|X_{\pi_i}). \tag{1}$$

Compared to tree models which are defined solely on observed variables (e.g., models obtained from the Chow-Liu algorithm (Chow and Liu, 1968)), latent tree models encompass a much larger classes of models, allowing more flexibility in modeling observed variables. This is evident if we compute the marginal distribution of the observed variables by summing out the latent ones,

$$p(\mathscr{O}) = \sum_{\mathscr{H}} \prod_{i=1}^{O+H} p(X_i|X_{\pi_i}). \tag{2}$$

This expression leads to complicated conditional independence structures between observed variables depending on the tree structure. In other words, latent tree models allow complex distributions over observed variables (e.g., clique models) to be expressed in terms of more tractable joint models over the augmented variable space. This leads to a significant saving in model parametrization.

## 3 Overview of Our Technique

To address nonparametric problems in tree-structured latent variable models, we will use a crucial technique called Hilbert space embedding of distributions and conditional distributions (an introduction is provided in Appendix A and B). The key idea is that the joint density of the observed variables under a tree-structured latent variable model can be expressed using a Hilbert space embedding $\mathcal{C}_{\mathscr{O}}$ which can be viewed as a higher order tensor (or multilinear operator). The density at point $x_1, \ldots, x_O$ can be evaluated by applying the multilinear operator on the feature-mapped data point, i.e., $\mathcal{C}_{\mathscr{O}} \bullet_1 \phi(x_1) \bullet_2 \ldots \bullet_O \phi(x_O)$. Under this view, the information about the density is fully captured by the Hilbert space embedding $\mathcal{C}_{\mathscr{O}}$.

Furthermore, the presence of the latent variables endows further structures in the Hilbert space embedding $\mathcal{C}_{\mathscr{O}}$. That is $\mathcal{C}_{\mathscr{O}}$ can be derived from a collection of simpler Hilbert of space embeddings $\{\mathcal{C}\}$ each of which involves only two or three observed variables. This decomposition allows us to design computationally efficient algorithms for learning both the structure and parameters associated with the latent variable model, and carry out efficient inference based on the decomposed representation.

One characteristic of our algorithms for learning structure and parameters is that they are based on the spectral property, especially low rank property, of the Hilbert space embedding $\mathcal{C}_{\mathscr{O}}$ due to the presence of latent variables. Our use of the spectral property of $\mathcal{C}_{\mathscr{O}}$ results in algorithms which are based on simple linear algebraic operation (or kernel matrix operation) and yet equipped with provable guarantee.

In the following, we will describe our techniques in 5 section. First, we will connect kernel density estimation with Hilbert space embedding of distributions. Second, we will explain the decomposed representation of the joint embedding $\mathcal{C}_{\mathscr{O}}$ of the tree-structured latent variable models. Third, we will use the decomposed form to design an efficient nonparametric inference algorithm. Fourth, we will derive our nonparametric parameter learning algorithm. Last, we will design the nonparametric structure learning algorithm using the spectral property of the embedding.

# 4 Kernel Density Estimator as Hilbert Space Embeddings

In this section, we connect traditional kernel density estimators (KDE) (Rosenblatt, 1956; Parzen, 1962; Silverman, 1986; Wasserman, 2006) with Hilbert space embeddings of distributions (Smola et al., 2007a). These two sets of methods are closely related to each other by the fact that both are trying to model non-Gaussian distribution in a nonparametric fashion, and kernel functions play a key role in both methods. The difference is that the studies of Hilbert space embeddings focus more on injectively representing distributions in a function space, while kernel density estimations concern more about estimating the actual values of the density. By building the connections, our purpose is to show that KDE can benefit from the view of Hilbert space embedding. In particular, we will later show that the Hilbert space embedding framework allows us to exploit KDE to model latent tree structures.

## 4.1 Kernel density estimator

The kernel density estimator (KDE), also called Parzen window estimator, is a nonparametric method for estimating the density function $p(x_1, \ldots, x_O)$ for a set of continuous random variables $\mathscr{O} = \{X_1, \ldots, X_O\}$. Given a dataset $\mathcal{D} = \{(x_1^i, \ldots, x_O^i)\}_{i=1}^n$ drawn i.i.d. from $p(x_1, \ldots, x_O)$, the KDE using product kernel is defined by

$$\widehat{p}(x_1, \ldots, x_O) = \frac{1}{n} \sum_{i=1}^{n} \prod_{j=1}^{O} K(x_j, x_j^i), \tag{3}$$

where $K(x, x')$ is a kernel function. A commonly used kernel function, which we will focus on, is the Gaussian RBF kernel $K(x, x') = \frac{1}{\sqrt{2\pi}\sigma} \exp(-\|x - x'\|^2 / 2\sigma^2)$. For Gaussian RBF kernel, there exists a feature map $\phi : \mathbb{R} \mapsto \mathcal{F}$ such that $K(x, x') = \langle \phi(x), \phi(x') \rangle_{\mathcal{F}}$, and the feature space has the reproducing property, i.e., for all $f \in \mathcal{F}$, $f(x) = \langle f, \phi(x) \rangle_{\mathcal{F}}$. Products of kernels are also kernels, which allows us to write $\prod_{j=1}^{O} K(x_j, x_j')$ as a single inner product $\left\langle \otimes_{j=1}^{O} \phi(x_j), \otimes_{j=1}^{O} \phi(x_j') \right\rangle_{\mathcal{F}^O}$. Here $\otimes_{j=1}^{O}(\cdot)$ denotes the outer product of $O$ feature vectors which results in a rank-1 tensor of order $O$, and the inner product can be understood by analogy with the finite dimensional case: given $x, y, z, x', y', z' \in \mathbb{R}^O$, $(x^\top x')(y^\top y')(z^\top z') = \langle x \otimes y \otimes z, \ x' \otimes y' \otimes z' \rangle_{\mathbb{R}^{d^3}}$.

## 4.2 A Hilbert space embedding view of KDE

To see the connection between KDE and Hilbert space embeddings of distributions, we compute the expected value of a KDE with respect to the random sample $\mathscr{S}$,

$$\mathbb{E}_{\mathcal{D}}[\widehat{p}(x_1, \ldots, x_O)] = \mathbb{E}_{\mathscr{O}} \left[ \prod_{j=1}^{O} K(x_j, X_j) \right] = \left\langle \mathbb{E}_{\mathscr{O}} \left[ \otimes_{j=1}^{O} \phi(X_j) \right], \otimes_{j=1}^{O} \phi(x_j) \right\rangle_{\mathcal{F}^O}, \tag{4}$$

where $\mathcal{C}_{\mathscr{O}} := \mathbb{E}_{\mathscr{O}} \left[ \otimes_{j=1}^{O} \phi(X_j) \right]$ is called the Hilbert space embedding of the density $p(x_1, \ldots, x_O)$ with tensor features $\otimes_{j=1}^{O} \phi(x_j)$ (Smola et al., 2007a). Furthermore, if we replace the embedding $\mathcal{C}_{\mathscr{O}}$ by its finite sample estimate $\widehat{\mathcal{C}}_{\mathscr{O}} := \frac{1}{n} \sum_{i=1}^{n} \left( \otimes_{j=1}^{O} \phi(x_j^i) \right)$, we recover the density estimator in (3).

## 4.3 Tensor expression for KDE

Using the tensor notation we can write equation (4) as

$$\left\langle \mathbb{E}_{\mathscr{O}}\left[\otimes_{j=1}^{O}\phi(X_j)\right], \otimes_{j=1}^{O}\phi(x_j)\right\rangle_{\mathcal{F}^O} = \mathcal{C}_{\mathscr{O}} \bullet_1 \phi(x_1) \bullet_2 \ldots \bullet_O \phi(x_O), \quad (5)$$

which means that the expected value of a KDE can be equivalently computed by first embedding the distributions into a tensor space and then multiply the embedding with the feature vectors of $x_1, \ldots, x_O$. We note that it is not easy for traditional KDE to exploit the fact that the embedding $\mathcal{C}_{\mathscr{O}}$ may have low rank structure due to the latent tree structure. In the next section, we will show that we can exploit the latent tree structure and come up with a factorized estimator for $\mathcal{C}_{\mathscr{O}}$.

## 5 Latent Tree Representation via Hilbert Space Embedding

When the conditional independent relation underlying a set of observed variables $\mathscr{O} = \{X_1, \ldots, X_O\}$ and hidden variables $\mathscr{H} = \{X_{O+1}, \ldots, X_{O+H}\}$ follows a latent tree structure, the joint embedding $\mathcal{C}_{\mathscr{O}}$ can factorize according to the tree structure. Instead of representing the density of the observed variables as a single Hilbert space embedding, we can represent it as a collection of Hilbert space embedding of conditional distributions, each of which involving only two variables. This is analogous to the factorization of a latent tree graphical model in equation (2), where we can represent such as a model using only the marginal distribution of the root node and the conditional distributions of variables with parent-child relations in the tree. Under a tree structure constraint, each variable has exactly one parent. In the nonparametric setting, we will instead use Hilbert space embedding of the marginal distribution and conditional distributions.

We will use the same kernel, $l(x, x') = \langle \psi(x), \psi(x') \rangle$ (with the associated RKHS $\mathcal{G}$), for all discrete latent variables. In particular, each latent variable $x$ can take up to $k$ distinct values $x \in \{1, \ldots, k\}$. The feature map $\psi(\cdot)$ is a mapping from $\{1, \ldots, k\}$ to $\{0, 1\}^k$ according to the rule that $\psi(x)$ is a length $k$ vector with all entries equal to 0 except the $x$th entry take value 1. For $x, x' \in \{1, \ldots, k\}$, their kernel function $l(x, x')$ is computed as the Euclidean inner product of $\psi(x)$ and $\psi(x')$.

Given the latent tree structure, we will choose an arbitrary latent variable as the root node $r$, Then we will represent the marginal distribution, $p(X_r)$, of the root node of the latent tree model as a Hilbert space embedding $\mathcal{C}_{r^3} := \mathbb{E}_{X_r}[\psi(x_r) \otimes \psi(x_r) \otimes \psi(x_r)]$. For each conditional distribution, $p(X_s|X_{\pi_s})$, we represent it as a conditional Hilbert embedding $\mathcal{C}_{s^2|\pi_s}\psi(X_{\pi_s}) := \mathbb{E}_{X_s|X_{\pi_s}}[\psi(X_s) \otimes \psi(X_s)]$ (if $X_s$ is a discrete latent variable) or $\mathcal{C}_{s|\pi_s}\psi(X_{\pi_s}) := \mathbb{E}_{X_s|X_{\pi_s}}[\phi(X_s)]$ (if $X_s$ is an observed continuous variable). With this representation, we will then be able to perform inference task on the latent tree graphical models, such as computing the marginal distribution of all observed variables by summing out all latent variables, or conditioning on the values of some observed variables to compute the distribution of other observed variables.

In the next section, we will first explain the inference (or query) task on a latent tree graphical model assuming that we are *given* the latent tree structures and the above mentioned conditional Hilbert space embedding representation of a latent tree graphical model. In this case, we can carry out the computation using a message passing algorithm. Next, we will discuss learning the Hilbert space embedding representation given information of the tree structure and the observed variables. Then, we will present our algorithm for discovering the tree structure of a latent variable model based on Hilbert space embedding of the joint distributions of pairs of observed variables.

# 6   Inference on Latent Tree Graphical Models

In this section, we assume the latent tree structure is known, i.e., for each variable $X_s$, we know its parent $X_{\pi_s}$. In addition, we assume all the conditional Hilbert space embedding of $p(X_s|X_{\pi_s})$ is given. Our goal is to conduct inference on the tree. For this, we can carry out the computation using a message passing algorithm.

We will focus on computing the expected KDE, $\mathbb{E}[\widehat{p}(x_1,\ldots,x_d)] = \mathbb{E}_{\mathscr{O}}\left[\prod_{j=1}^{O} K(x_j, X_j)\right]$, for the observed variables in a latent tree graphical model. We will show that the computation can be carried out based on the representation we discussed in the last section. More specifically, in a tree graphical model, we can carry out the computation efficiently via the message passing algorithm (Pearl, 2001):

- At a leaf node (corresponding to an observed variable), we pass the following message to its parent $m_s(\cdot) = \mathbb{E}_{X_s|X_{\pi_s}=\cdot}[K(x_s, X_s)]$.

- An internal latent variable aggregrates incoming messages from its two children and then sends an outgoing message to its own parent $m_s(\cdot) = \mathbb{E}_{X_s|X_{\pi_s}=\cdot}[m_{\iota_s}(X_s)m_{\rho_s}(X_s)]$.

- Finally, at the root node, all incoming messages are multiplied together and the root variable is integrated out $b_r = \mathbb{E}_{\mathscr{O}}[\prod_{j=1}^{O} K(x_j, X_j)] = \mathbb{E}_{X_r}[m_{\iota_s}(X_r)m_{\rho_s}(X_r)m_{\omega_r}(X_r)]$.

All the above 3 message update operations can be expressed by the corresponding Hilbert space embeddings. Following Song et al. (2010) on the work of nonparametric tree graphical models, we can thus express message passing algorithm as linear operations in the feature space,

- At a leaf node, we have $m_{ts}(\cdot) = \mathbb{E}_{X_s|X_{\pi_s}=\cdot}[K(x_s, X_s)] = \mathcal{C}_{s|\pi_s} \bullet_1 \phi(x_s)$.

- At internal nodes, we use a tensor product reproducing kernel Hilbert space $\mathcal{G}^2 := \mathcal{G} \otimes \mathcal{G}$, under which the product of incoming messages can be written as a single inner product,

$$m_{\iota_s}(X_s)\, m_{\rho_s}(X_s) = \langle m_{\iota_s}, \psi(X_s)\rangle_{\mathcal{G}} \langle m_{\rho_s}, \psi(X_s)\rangle_{\mathcal{G}} = \langle m_{\iota_s} \otimes m_{\rho_s}, \psi(X_s) \otimes \phi(X_s)\rangle_{\mathcal{G}^2}.$$

  Then the message update becomes

$$m_s(\cdot) = \left\langle m_{\iota_s} \otimes m_{\rho_s},\ \mathbb{E}_{X_s|X_{\pi_s}=\cdot}\left[\phi(X_s) \otimes \phi(X_s)\right]\right\rangle_{\mathcal{G}^2} = \mathcal{C}_{s^2|\pi_s} \bullet_1 m_{\iota_s} \bullet_2 m_{\rho_s}, \tag{6}$$

- Finally, at the root nodes, we use the property of tensor product features $\mathcal{G}^3 := \mathcal{G} \otimes \mathcal{G} \otimes \mathcal{G}$ and arrive at:

$$\mathbb{E}_r[m_{\iota_r}(X_r)\, m_{\rho_r}(X_r)\, m_{\omega_r}(X_r)] = \langle m_{\iota_r} \otimes m_{\rho_r} \otimes m_{\omega_r}, \mathbb{E}_{X_r}[\phi(X_r) \otimes \phi(X_r) \otimes \phi(X_r)]\rangle_{\mathcal{G}^3}$$
$$= \mathcal{C}_{r^3} \bullet_1 m_{\iota_r} \bullet_2 m_{\rho_r} \bullet_3 m_{\omega_r}. \tag{7}$$

We note that traditional kernel density estimator needs to estimate a tensor of order $O$ involving all observed variables (i.e., Equation (5)). By exploiting the conditional independence structure of latent tree models, we only need to estimate tensors of much smaller orders. In particular, we only need to estimate tensors involving two variables (for each parent-child pair), then the density can be estimated via message passing algorithms using these tensors of much smaller order.

The above inference task for latent tree graphical models with Hilbert space embedding representation requires the exact knowledge of conditional embedding operators associated with latent variables. Next we will show that given information only on the tree structure and the observed variables, we can recover a transformed representation of the conditional embedding operators and perform the same inference task.

# 7 Parameter Estimation

In this section, we assume the latent tree structure is known, i.e., for each variable $X_s$, we know its parent $X_{\pi_s}$. Our goal is to estimate all the conditional embedding operators associated with $p(X_s|X_{\pi_s})$.

From (6) and (7), our key observation is that if we can recover the conditional embedding operators up to some invertible transformations, we will still be able to obtain the same final results for the message passing algorithm. For example, for each leaf node $X_s$, we define an invertible transformation $T_s \in \mathbb{R}^{k \times k}$, where the subscript is used to indicate that the transformation $T_s$ is specific to node $X_s$. If we change to a different node $X_{s'}$, then the transformation $T_{s'}$ will also change accordingly. Then we can transform the message, $m_s$, from a leaf node as

$$\widetilde{m}_s = (\mathcal{C}_{s|\pi_s} \times_2 T_s) \bullet_1 \phi(x_s) = T_s \mathcal{C}_{s|\pi_s}^\top \phi(x_s), \tag{8}$$

where the tensor-matrix product notation $\mathcal{C} \times_i T$ means that we multiply the $i$-th mode of $\mathcal{C}$ with the columns of $T$ (Kolda and Bader, 2009). Then in an internal node, we will introduce three invertible transformations $T_s, T_{\iota_s}, T_{\rho_s} \in \mathbb{R}^{k \times k}$. We note that in this case, we do not have complete freedom in choosing the invertible transformations: $T_{\iota_s}$ and $T_{\rho_s}$ are determined by the transformation chosen by the child node $X_{\iota_s}$ and $X_{\rho_s}$ of the current node $X_s$. Then the outgoing message, $m_s$, from this internal node becomes

$$\widetilde{m}_s = (\mathcal{C}_{s^2|\pi_s} \times_1 T_{\iota_s}^{-1} \times_2 T_{\rho_s}^{-1} \times_3 T_s) \bullet_1 \widetilde{m}_{\iota_s} \bullet_2 \widetilde{m}_{\rho_s} \tag{9}$$

where we have defined $\widetilde{m}_{\iota_s} = T_{\iota_s} m_{\iota_s}$ and $\widetilde{m}_{\rho_s} = T_{\rho_s} m_{\rho_s}$. Finally, at the root node, we also introduce three invertible transformations $T_{\iota_s}, T_{\rho_s}, T_{\omega_s} \in \mathbb{R}^{k \times k}$ which are determined by the choice of transformations from its three child nodes. Then we obtain the final result

$$b_r = (\mathcal{C}_{r^3} \times_1 T_{\iota_s}^{-1} \times_2 T_{\rho_s}^{-1} \times_3 T_{\omega_r}^{-1}) \bullet_1 \widetilde{m}_{\iota_s} \bullet_2 \widetilde{m}_{\rho_s} \bullet_3 \widetilde{m}_{\omega_s}, \tag{10}$$

where $\widetilde{m}_{\omega_s} = T_{\omega_s} m_{\omega_s}$. Basically, all the invertible transformations $T$'s cancel out with each other, and the final result $b_r$ remains unchanged. However, these transformations provide us additional degrees of freedom for algorithm design: We can more carefully choose the invertible transforms so that the transformed representation can be recovered from observed quantities without the need for accessing the latent variables.

We will show that these transformations $T$'s can be constructed from singular vectors $U$ of cross covariance operator of certain pairs of observed variables. To illustrate the key ideas, we consider the simple case for the transformed message $\widetilde{m}_s$ from the leaf node (equation 8). Let $U_{s\pi_s^*}$ be the left singular vectors of $\mathcal{C}_{s\pi_s^*}$ corresponding to the top $k$ singular values. We use the transformation $T_s = (\mathcal{C}_{s|\pi_s}^\top U_{s\pi_s^*})^{-1}$, then we have that

$$\widetilde{m}_s = (\mathcal{C}_{s|\pi_s} \times_2 (\mathcal{C}_{s|\pi_s}^\top U_{s\pi_s^*})^{-1}) \bullet_1 \phi(x_s) = (\mathcal{C}_{s|\pi_s}^\top U_{s\pi_s^*})^{-1} \mathcal{C}_{s|\pi_s}^\top \phi(x_s). \tag{11}$$

Then, to remove the dependency of the expression on latent variables, we multiply both sides by $\mathcal{C}_{\alpha_s s} U_{s\pi_s^*}$ and obtain

$$\begin{aligned}
\mathcal{C}_{\alpha_s s} U_{s\pi_s^*} \widetilde{m}_s &= \mathcal{C}_{\alpha_s s} U_{s\pi_s^*} (\mathcal{C}_{s|\pi_s}^\top U_{s\pi_s^*})^{-1} \mathcal{C}_{s|\pi_s}^\top \phi(x_s) \\
&= \mathcal{C}_{\alpha_s|\pi_s} \mathcal{C}_{\pi_s \pi_s} \mathcal{C}_{s|\pi_s}^\top U_{s\pi_s^*} (\mathcal{C}_{s|\pi_s}^\top U_{s\pi_s^*})^{-1} \mathcal{C}_{s|\pi_s}^\top \phi(x_s) \\
&= \mathcal{C}_{\alpha_s|\pi_s} \mathcal{C}_{\pi_s \pi_s} \mathcal{C}_{s|\pi_s}^\top \phi(x_s) \\
&= \mathcal{C}_{\alpha_s s} \phi(x_s) \tag{12}
\end{aligned}$$

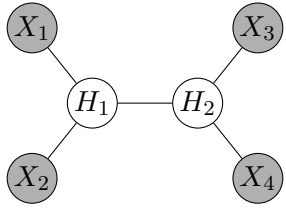

Figure 2: A tree-structured latent variable model with 4 observed variable $X_1, X_2, X_3, X_4$, connected via two hidden variables $H_1, H_2$.

where in the second and third equalities we have used Hilbert space embedding expression for the relation $p(X_{\alpha_s}, X_s) = \sum_{x_{\pi_s}} p(X_{\alpha_s}|x_{\pi_s}) p(x_{\pi_s}) p(X_s|x_{\pi_s})$, i.e., $\mathcal{C}_{\alpha_s s} = \mathcal{C}_{\alpha_s|\pi_s} \mathcal{C}_{\pi_s \pi_s} \mathcal{C}_{s|\pi_s}^\top$. Finally, based on the derivation in equation (12), we have that

$$\widetilde{m}_s = (\mathcal{C}_{\alpha_s s} U_{s\pi_s^*})^\dagger \mathcal{C}_{\alpha_s s} \phi(x_s), \tag{13}$$

which depends only on information of observed variables. The general pattern of this derivation is that we can relate the transformed latent quantity to the observed quantities by choosing appropriate invertible transformations. Similar strategy can be appled to $\widetilde{\mathcal{C}}_{s^2|\pi_s} := \mathcal{C}_{s^2|\pi_s} \times_1 T_{\iota_s}^{-1} \times_2 T_{\rho_s}^{-1} \times_3 T_s$ in the internal message update, and to $\widetilde{\mathcal{C}}_{r^3} := \mathcal{C}_{r^3} \times_1 T_{\iota_s}^{-1} \times_2 T_{\rho_s}^{-1} \times_3 T_{\omega_r}^{-1}$ in the update at the root. To summarize these results, we state the following theorem:

**Theorem 1.** *The transformed quantities involving latent variables can be computed via observed quantities using the following formuli*

- *For observed variables, $\widetilde{m}_s = (\mathcal{C}_{\alpha_s s} U_{s\pi_s^*})^\dagger \mathcal{C}_{\alpha_s s} \phi(x_s)$.*
- *For latent variables, $\widetilde{\mathcal{C}}_{s^2|\pi_s} = \mathcal{C}_{\iota_s^* \rho_s^* \pi_s^*} \times_1 U_{\iota_s^* \pi_s^*}^\top \times_2 U_{\rho_s^* \pi_s^*}^\top \times_3 (\mathcal{C}_{\pi_s^* \iota_s^*} U_{\iota_s^* \pi_s^*})^\dagger$.*
- *For the root node, $\widetilde{\mathcal{C}}_{r^3} = \mathcal{C}_{\iota_s^* \rho_s^* \omega_s^*} \times_1 U_{\iota_s^* \rho_s^*}^\top \times_2 U_{\rho_s^* \omega_s^*}^\top \times_3 U_{\omega_s^* \iota_s^*}^\top$.*

*Proof.* **Sketch:** Similar to the derivation in (Parikh et al., 2011), we replace all conditional distributions by conditional embedding operators and get the desired result. □

### 7.1 Computation

When we use the factorization of Hilbert space embeddings from Theorem 1 for density estimation, it leads us to a very efficient algorithm for computing the expected kernel density $\mathbb{E}\widehat{p}(x_1, \ldots, x_O) = \mathbb{E}_\mathscr{O}\left[\prod_{j=1}^O K(x_j, X_j)\right]$. The main computational cost only involves a sequence of singular value decompositions of pairwise cross-covariance operators. Once the transformed quantities are obtained, we can then use them in the message passing algorithm to obtain the final belief.

Given a sample $\mathscr{D} = \{(x_1^i, \ldots, x_O^i)\}_{i=1}^n$ drawn i.i.d. from $p(x_1, \ldots, x_O)$, the spectral algorithm for latent tree graphical models proceeds by first performing a "thin" SVD of the sample covariance operators. For instance, for two variables $X_s$ and $X_t$, we denote the feature matrices by $\Upsilon =$ $(\phi(x_s^1), \ldots, \phi(x_s^n))$ and $\Phi = (\phi(x_t^1), \ldots, \phi(x_t^n))$, and estimate $\widehat{\mathcal{C}}_{ts} = \frac{1}{n}\Phi\Upsilon^\top$. Then the left singular vector $v = \Phi\alpha$ ($\alpha \in \mathbb{R}^n$) can be estimated as follows

$$\Phi\Upsilon^\top\Upsilon\Phi^\top v = \beta v \Leftrightarrow \mathbf{LKL}\alpha = \beta\mathbf{L}\alpha \quad (\beta \in \mathbb{R}), \tag{14}$$

where $\mathbf{K} = \Upsilon^\top\Upsilon$ and $\mathbf{L} = \Phi^\top\Phi$ are the kernel matrices, and $\alpha$ is the generalized eigenvector. After normalization, we have $v = \Phi\alpha/\sqrt{\alpha^\top\mathbf{L}\alpha}$. Then the $U$ is the column concatenation of the top $k$ left singular vectors, i.e., $\widehat{U} = (v_1, \ldots, v_k)$. In practice, fast computation of the kernel SVD can be carried out by first performing an incomplete Cholesky decomposition of the kernel matrices (Shawe-Taylor and Cristianini, 2004). If we let $A := (\alpha_1, \ldots, \alpha_k) \in \mathbb{R}^{n \times k}$ be the column concatenation of the $k$ top $\alpha_i$, and $D := \text{diag}\left((\alpha_1^\top\mathbf{L}\alpha_1)^{-1/2}, \ldots, (\alpha_k^\top\mathbf{L}\alpha_k)^{-1/2}\right) \in \mathbb{R}^{k \times k}$, we can concisely express $\widehat{U} = \Phi AD$.

When the number of samples is large, we will use incomplete Cholesky decomposition for kernel matrices to further speed up the computation. In this case the kernel matrix $\mathbf{K}$ for samples from a variable $X$ can be factorized as $\mathbf{K} = R_X^\top R_X$, where $R_X \in \mathbb{R}^{r \times n}$ comes from Cholesky decomposition with $r \ll n$. Basically, with incomplete Cholesky decomposition, everything goes back to finite dimensional operations.

## 7.2 Example

We can factorize the embedding of the marginal distributions of the four observed variables in the latent tree structure in Figure 2 as

$$\begin{aligned}\mathcal{C}_{X_1X_2X_3X_4} &= \mathbb{E}_{X_1X_2X_3X_4}[\phi(X_1) \otimes \phi(X_2) \otimes \phi(X_3) \otimes \phi(X_4)] \\ &= (\mathcal{C}_{H_1^3} \times_1 \mathcal{C}_{X_1|H_1} \times_2 \mathcal{C}_{X_2|H_2}) \times_3 (\mathcal{C}_{H_2^2|H_1} \times_1 \mathcal{C}_{X_3|H_2} \times_2 \mathcal{C}_{X_4|H_2}).\end{aligned}$$

Then the term $\mathcal{C}_{H_1^3}$ can be transformed as

$$\begin{aligned}\mathcal{C}_{H_1^3} \times_1 S_{X_1} \times_2 S_{X_2} \times_3 S_{X_3} &= \mathcal{C}_{H_1^3} \times_1 (U_{X_1}^\top \mathcal{C}_{X_1|H_1}) \times_2 (U_{X_2}^\top \mathcal{C}_{X_2|H_1}) \times_3 (U_{X_3|H_1}^\top \mathcal{C}_{X_3|H_1}) \\ &= \mathcal{C}_{X_1X_2X_3} \times_1 U_{X_1}^\top \times_2 U_{X_2}^\top \times_3 U_{X_3}^\top.\end{aligned}$$

The term $\mathcal{C}_{H_2^2|H_1}$ can be transformed as

$$\begin{aligned}\mathcal{C}_{H_2^2|H_1} &\times_3 (\mathcal{C}_{X_1|H_1}\mathcal{C}_{H_1^2}S_{X_3}^\top S_{X_3}^{-\top}) \times_1 S_{X_3} \times_2 S_{X_4} \\ &= \mathcal{C}_{H_2^2|H_1} \times_3 (\mathcal{C}_{X_1X_3}U_{X_3}S_{X_3}^{-\top}) \times_1 (U_{X_3}^\top \mathcal{C}_{X_3|H_2}) \times_2 (U_{X_4}^\top \mathcal{C}_{X_4|H_2}) \\ &= \mathcal{C}_{H_2^2|H_1} \times_3 (\mathcal{C}_{X_1X_3}U_{X_3}S_{X_3}^{-\top}) \times_1 (U_{X_3}^\top \mathcal{C}_{X_3|H_2}) \times_2 (U_{X_4}^\top \mathcal{C}_{X_4|H_2}) \\ &= \mathcal{C}_{X_3X_4X_1} \times_1 U_{X_3}^\top \times_2 U_{X_4}^\top.\end{aligned}$$

We therefore have

$$\mathcal{C}_{H_2^2|H_1} \times_3 S_{X_3}^{-\top} \times_1 S_{X_3} \times_2 S_{X_4} = \mathcal{C}_{X_3X_4X_1} \times_1 U_{X_3}^\top \times_2 U_{X_4}^\top \times_3 (\mathcal{C}_{X_1X_3}U_{X_3})^\dagger$$

and the term $\mathcal{C}_{X_1|H_1}$ can be transformed as

$$\mathcal{C}_{X_1|H_1}S_{X_1}^{-1} = \mathcal{C}_{X_1X_2}(U_{X_1}^\top \mathcal{C}_{X_1X_2})^\dagger.$$

Then the computation of $U$ using the approximated covariance matrix can be carried out as follows

$$\frac{1}{n^2}(R_{X_1}R_{X_2}^\top)(R_{X_2}R_{X_1}^\top)U_{X_1} = U_{X_1}\Gamma_1 \tag{15}$$

$$\frac{1}{n^2}(R_{X_2}R_{X_1}^\top)(R_{X_1}R_{X_2}^\top)U_{X_2} = U_{X_2}\Gamma_2 \tag{16}$$

$$\frac{1}{n^2}(R_{X_3}R_{X_1}^\top)(R_{X_1}R_{X_3}^\top)U_{X_3} = U_{X_3}\Gamma_3 \tag{17}$$

$$\frac{1}{n^2}(R_{X_4}R_{X_1}^\top)(R_{X_1}R_{X_4}^\top)U_{X_4} = U_{X_4}\Gamma_4, \tag{18}$$

where $\Gamma$ is a diagonal matrix of the corresponding eigenvalues. Furthermore, the transformed parameters of the latent tree graphical model can also be obtained from the Cholesky factor. For example

$$\mathcal{C}_{X_1X_2}(U_{X_1}^\top \mathcal{C}_{X_1X_2})^\dagger \approx \frac{1}{n}R_{X_1}R_{X_2}^\top (U_{X_1}^\top \frac{1}{n}R_{X_1}R_{X_2}^\top)^\dagger. \tag{19}$$

To evaluate a new point, we first obtain the incomplete Cholesky decomposition for the new point $r_{X_1}, r_{X_2}, r_{X_3}, r_{X_4}$. Then we obtain the density estimation by

$$\mathcal{C}_{X_1X_2X_3X_4} \bullet_1 r_{X_1} \bullet_2 r_{X_2} \bullet_3 r_{X_3} \bullet_4 r_{X_4} \tag{20}$$

which can be computed by message passing algorithm.

## 8 Structure Learning

In the last section we focuses on estimating the Hilbert space embedding under the assumption that the structure of the latent tree is known. In this section, we focus on learning latent tree structure based observational data. We exploit a distance-based method for iteratively constructing latent trees which takes a distance matrix between all pairs of observed variables as input and outputs an estimated tree structure by iteratively adding hidden nodes.We provide theoretical guarantees of the structure learning algorithm. We start with some definitions.

### 8.1 Tree Metric and Pseudo-determinant for Non-Gaussian Variables

One concept that plays a pivotal role in our latent tree structure learning algorithm is tree metric. More specifically, if the joint probability distribution $p(\mathcal{X})$ has a latent tree structure, then a distance measure $d_{st}$ between an arbitrary pair of variables $X_s$ and $X_t$ is called tree metric if it satisfies the path additive condition: $d_{st} = \sum_{(u,v)\in Path(s,t)} d_{uv}$. The formal definition of tree metric is as follows.

**Definition 2** (Tree Metric). *Let $\mathcal{V}$ be a set of nodes. A metric $(\mathcal{V}, d)$ is a tree metric if there is a tree $\mathcal{T}$ with non-negative edge lengths whose nodes contain $\mathcal{V}$ such that for every $s, t \in \mathcal{V}$, we have that $d_{st}$ is equal to the sum of the lengths of the edges on the unique $(s, t)$ path in the tree $\mathcal{T}$. Tree metrics are also called additive metrics.*

For discrete and Gaussian variables, tree metric can be defined via the determinant $|\cdot|$

$$d_{st} = -\log|\mathcal{C}_{st}| + \tfrac{1}{2}\log|\mathcal{C}_{ss}| + \tfrac{1}{2}\log|\mathcal{C}_{tt}|, \tag{21}$$

where $\mathcal{C}_{st}$ is the joint probability matrix in the discrete case and the covariance matrix in the Gaussian case; $\mathcal{C}_{ss}$ is the diagonalized marginal probability matrix in the discrete case and marginal variance matrix in the Gaussian case. However, the definition of tree metric in (21) is restrictive since it requires all discrete variables to have the same number of states and all Gaussian variables to have the same dimension. For our case where the observed variables are continuous and non-Gaussian. The problem with the above definition is that determinant is only defined for square and non-singular matrices. Therefore, the definition in (21) is not suitable. To overcome this drawback, we define a tree metric based on pseudo-determinant which works for more general operators. First, we list an extra assumption.

(A2) Let $\sigma_k(C_{st})$ be the $k$th singular value of $\mathcal{C}$, we assume that $\sigma_{\min} := \inf_{s,t} \sigma_k(C_{st}) > 0$.

Under the condition that the value $k$ in Assumption (A1) is known, we define the pseudo-determinant of a covariance operator $\mathcal{C}$ as the product of all its non-zero singular values, i.e.,

$$|\mathcal{C}|_* = \prod_{i=1}^{k} \sigma_i(\mathcal{C}).$$

We then generalize the definition of tree metric in (21) from the covariance matrix setting to the more general covariance operator setting. In particular, we define a nonparametric distance metric between two variables $s$ and $t$ as

$$d_{st} = -\log |\mathcal{C}_{st}|_* + \tfrac{1}{2} \log |\mathcal{C}_{ss}|_* + \tfrac{1}{2} \log |\mathcal{C}_{tt}|_*. \tag{22}$$

The following theorem shows that (22) is a valid tree metric.

**Theorem 3.** *The distance define in equation* (22) *is a tree metric.*

*Proof.* We prove this by induction on the path length. We first show that the additive property holds for a path $X_s - X_u - X_t$ which only involves a single hidden variable $X_u$. For this, we exploit the relationship between eigenvalues and singular values and have

$$d_{st} = -\tfrac{1}{2} \log |\mathcal{C}_{st}\mathcal{C}_{st}^\top|_* + \tfrac{1}{4} \log |\mathcal{C}_{ss}\mathcal{C}_{ss}^\top|_* + \tfrac{1}{4} \log |\mathcal{C}_{tt}\mathcal{C}_{tt}^\top|. \tag{23}$$

Furthermore, using the Markov property, we factorize $|\mathcal{C}_{st}\mathcal{C}_{st}^\top|_*$ into $|\mathcal{C}_{s|u}\mathcal{C}_{uu}\mathcal{C}_{t|u}^\top \mathcal{C}_{t|u}\mathcal{C}_{uu}\mathcal{C}_{s|u}^\top|_*$. According to the Sylvester's determinant theorem, the latter is equal to $|\mathcal{C}_{s|u}^\top \mathcal{C}_{s|u}\mathcal{C}_{uu}\mathcal{C}_{t|u}^\top \mathcal{C}_{t|u}\mathcal{C}_{uu}|_*$ by flipping $\mathcal{C}_{s|u}^\top$ to the front. By introducing two copies of $|\mathcal{C}_{uu}|$ and rearraging the terms, we have

$$|\mathcal{C}_{st}\mathcal{C}_{st}^\top|_* = \frac{|\mathcal{C}_{s|u}\mathcal{C}_{uu}\mathcal{C}_{uu}\mathcal{C}_{s|u}^\top|_* |\mathcal{C}_{t|u}\mathcal{C}_{uu}\mathcal{C}_{uu}\mathcal{C}_{t|u}^\top|_*}{|\mathcal{C}_{uu}||\mathcal{C}_{uu}|} = \frac{|\mathcal{C}_{su}\mathcal{C}_{su}^\top|_* |\mathcal{C}_{tu}\mathcal{C}_{tu}^\top|_*}{|\mathcal{C}_{uu}\mathcal{C}_{uu}|_*}. \tag{24}$$

We then plug this into (23) and get

$$d_{st} = -\tfrac{1}{2} \log |\mathcal{C}_{su}\mathcal{C}_{su}^\top|_* - \tfrac{1}{2} \log |\mathcal{C}_{tu}\mathcal{C}_{tu}^\top|_* + \tfrac{1}{2} \log |\mathcal{C}_{uu}\mathcal{C}_{uu}^\top|_* + \tfrac{1}{4} \log |\mathcal{C}_{ss}\mathcal{C}_{ss}^\top|_* + \tfrac{1}{4} \log |\mathcal{C}_{tt}\mathcal{C}_{tt}^\top|$$
$$= d_{su} + d_{ut}.$$

The proof of the general cases follows similar arguments. □

## 8.2 Emiprical Tree Metric Estimator and Its Concentration Property

The definition in (22) involves population quantities. Given observational data, the pseudo-determinant tree distance between two variables $s$ and $t$ can be estimated by the following plug-in estimator

$$\widehat{d}_{st} = -\sum_{i=1}^{k} \log\left[\sigma_i(\widehat{\mathcal{C}}_{st})\right] + \frac{1}{2}\sum_{i=1}^{k} \log\left[\sigma_i(\widehat{\mathcal{C}}_{ss})\right] + \frac{1}{2}\sum_{i=1}^{k} \log\left[\sigma_i(\widehat{\mathcal{C}}_{tt})\right], \tag{25}$$

where $\widehat{\mathcal{C}}$ is an estimator of the covariance operator. From the definition in (25), we see that the key to evaluate the tree metric is to calculate $\sigma_i(\widehat{\mathcal{C}}_{st})$ and $\sigma_i(\widehat{\mathcal{C}}_{ss})$. For this, we use the same generalized eigenvalue decomposition method as described in Section 7.1.

In this subsection, we show that the the estimated distance $\widehat{d}_{st}$ converges to its popular quantity $d_{st}$ with a fast rate of convergence. More specifically, let $K(x, x')$ be a universal reproducing kernel satisfying

$$\sup_{x,x'} K(x, x') \leq \kappa.$$

We define

$$T_{\mathcal{F}}^{(ss)} = \int_{\Omega} \langle \cdot, K(x_s, \cdot) \rangle_{\mathcal{F}} K(x_s, \cdot) dP(x_s) = \mathcal{C}_{ss}, \tag{26}$$

$$T_n^{(ss)} = \frac{1}{n}\sum_{i=1}^{n} \langle \cdot, K(x_s^i, \cdot) \rangle_{\mathcal{F}} K(x_s^i, \cdot) = \widehat{\mathcal{C}}_{ss}, \tag{27}$$

and

$$T_{\mathcal{F}}^{(st)} = \int_{\Omega \times \Omega} \langle \cdot, K(x_t, \cdot) \rangle_{\mathcal{F}} K(x_s, \cdot) dP(x_s, x_t) = \mathcal{C}_{st}, \tag{28}$$

$$T_n^{(st)} = \frac{1}{n}\sum_{i=1}^{n} \langle \cdot, K(x_t^i, \cdot) \rangle_{\mathcal{F}} K(x_s^i, \cdot) = \widehat{\mathcal{C}}_{st}. \tag{29}$$

The following lemmas from Rosasco et al. (2010) show that $T_{\mathcal{F}}^{(ss)}$ and $T_{\mathcal{F}}^{(st)}$ have the same set of eigenvalues (up to possible zero entries ) as $T_n^{(ss)}$ and $T_n^{(st)}$.

**Lemma 4** (Rosasco et al. (2010) ). $T_{\mathcal{F}}^{(ss)}$ and $T_{\mathcal{F}}^{(st)}$ have the same set of eigenvalues (up to possible zero entries) as $T_n^{(ss)}$ and $T_n^{(st)}$ .

The following theorem is proved by Rosasco et al. (2010)

**Theorem 5.** Let $\delta \in (0, 1)$ and $\|\cdot\|_{HS}$ be the Hilbert-Schdmit norm, we have

$$\mathbb{P}\left(\left\|T_{\mathcal{F}}^{(ss)} - T_n^{(ss)}\right\|_{HS} \leq \frac{2\sqrt{2}\kappa\delta}{\sqrt{n}}\right) \geq 1 - 2\exp(-\delta). \tag{30}$$

$$\mathbb{P}\left(\left\|T_{\mathcal{F}}^{(st)} - T_n^{(st)}\right\|_{HS} \leq \frac{2\sqrt{2}\kappa\delta}{\sqrt{n}}\right) \geq 1 - 2\exp(-\delta). \tag{31}$$

*Proof.* The result in (30) has been shown by Rosasco et al. (2010). We only need to show (31). Let $\xi_i$ be defined as

$$\xi_i = \langle \cdot, K(x_t^i, \cdot) \rangle_{\mathcal{F}} K(x_s^i, \cdot) - T_{\mathcal{H}}^{(st)}. \tag{32}$$

It is easy to see that $\mathbb{E}\xi_i = 0$. Furthermore, we have

$$\sup_{s,t} \|\langle \cdot, K(x_t, \cdot) \rangle_{\mathcal{F}} K(x_s, \cdot)\|_{HS}^2 \leq \kappa^2, \tag{33}$$

which implies that $\left\|T_{\mathcal{F}}^{(st)}\right\|_{HS} \leq 2\kappa$. The result then follows from the Hoeffding's inequality in Hilbert space. □

The next theorem provides a uniform concentration result of the estimated tree metric $\hat{d}_{st}$ to its population quantity.

**Theorem 6.** *Under Assumption (A3), we have*

$$\mathbb{P}\left(\sup_{st} \left|\hat{d}_{st} - d_{st}\right| \leq \frac{8\kappa}{\sigma_{\min}} \cdot \sqrt{\frac{2k \log |\mathcal{O}|}{n}}\right) \geq 1 - o(1). \tag{34}$$

*Proof.* Recall from Assumption (A2) that $\sigma_{\min} = \inf_{s,t} \sigma_k(\mathcal{C}_{st})$, there exists a finite $n_0$, for $n \geq n_0$, we have, for $1 \leq i \leq k$,

$$\frac{1}{2} \leq \inf_{s,t} \frac{\sigma_i(\widehat{\mathcal{C}}_{st})}{\sigma_i(\mathcal{C}_{st})} \leq \sup_{s,t} \frac{\sigma_i(\widehat{\mathcal{C}}_{st})}{\sigma_i(\mathcal{C}_{st})} \leq \frac{3}{2}.$$

We then have

$$\left|\sum_{i=1}^k \log[\sigma_i(\widehat{\mathcal{C}}_{st})] - \sum_{i=1}^k \log[\sigma_i(\mathcal{C}_{st})]\right| \leq \sum_{i=1}^k \left|\log\left[\frac{\sigma_i(\widehat{\mathcal{C}}_{st}) - \sigma_i(\mathcal{C}_{tt}))}{\sigma_i(\mathcal{C}_{st})} + 1\right]\right|$$

$$\leq 2 \sum_{i=1}^k \frac{\left|\sigma_i(\widehat{\mathcal{C}}_{st}) - \sigma_i(\mathcal{C}_{st})\right|}{\sigma_i(\mathcal{C}_{st})}$$

$$\leq \frac{2}{\sigma_{\min}} \sum_{i=1}^k \left|\sigma_i(\widehat{\mathcal{C}}_{st}) - \sigma_i(\mathcal{C}_{st})\right|$$

$$\leq \frac{2\sqrt{k}}{\sigma_{\min}} \sqrt{\sum_{i=1}^k \left[\sigma_i(\widehat{\mathcal{C}}_{st}) - \sigma_i(\mathcal{C}_{st})\right]^2}$$

$$\leq \frac{2\sqrt{k}}{\sigma_{\min}} \left\|T_{\mathcal{F}}^{(st)} - T_n^{(st)}\right\|_{HS}$$

$$\leq \frac{\sqrt{k}}{\sigma_{\min}} \frac{4\sqrt{2}\kappa\sqrt{\delta}}{\sqrt{n}}$$

with probability larger than $1 - 2\exp(-\delta)$. Here the second to last inequality follows from the Mirsky Theorem. Using the above arguments and the union bound, we get the desired result. □

## 8.3 Structure Learning Algorithm

Once we define the tree metric as in the previous section, we can the neighbor-joint algorithm from the phylogeny tree literature to recover the latent tree structure (Saitou et al., 1987). The neighbor joint algorithm takes a distance matrix between any pair of observed variables as input and output a tree by iteratively adding hidden nodes. More specifically, let $s$ and $t$ be two nodes in the latent tree structure. They could be either observed or latent variables. We call $d_{st}$ to be the information distance between $s$ and $t$. The neighbor joining algorithm requires the input of the distance matrix between all observed variables. We summarize the algorithm in Algorithm 1.

---
**Algorithm 1** Neighbor Joining Algorithm
---
**Input:** Pairwise distance $d_{st}$ between observed variables in $\mathscr{O}$
**Output:** Latent tree structure $\mathscr{T} = (\mathscr{V}, \mathscr{E})$
  Initialize the latent tree structure: $\mathscr{V} = \{1, ..., |\mathscr{O}|\}, \mathscr{E} = \emptyset$
  Define a working set: $\mathscr{N} = \{1, ..., |\mathscr{O}|\}$
  $u = |\mathscr{O}| + 1$
  **while** $|\mathscr{N}| \geq 3$ **do**
    **for** $s \in \mathscr{N}$ **do**
      **for** $t \in \mathscr{N}, t \neq s$ **do**
        $Q_{st} = Q_{ts} = (|\mathscr{N}| - 2)d_{st} - \sum_{l=1}^{|\mathscr{N}|} d_{sl} - \sum_{l=1}^{|\mathscr{N}|} d_{tl}$
      **end for**
    **end for**
  $(s^*, t^*) = \operatorname{argmin}_{s,t \in \mathscr{N}, s \neq t} Q_{st}$
  Create a new node with index $u$ to join node $s^*$ and $t^*$
  Update distances to the new node $u$:

$$d_{s^*u} = d_{us^*} = \frac{1}{2}d_{s^*t^*} + \frac{1}{2(|\mathscr{N}|-2)}\left(\sum_{l=1}^{|\mathscr{N}|} d_{s^*l} - \sum_{l=1}^{|\mathscr{N}|} d_{t^*l}\right),$$

$$d_{t^*u} = d_{ut^*} = d_{s^*u} - d_{s^*u}$$

$$d_{lu} = d_{ul} = \frac{1}{2}\left(d_{s^*l} + d_{t^*l} - d_{s^*t^*}\right), \forall l \in \mathscr{N}, l \neq s^*, l \neq t^*$$

  Update the working set:

$$\mathscr{N} = \mathscr{N} \setminus \{s^*, t^*\}, \quad \mathscr{N} = \mathscr{N} \cup \{u\}, \quad u = u + 1$$

  Update latent tree structure $\mathscr{T}$:

$$\mathscr{V} = \mathscr{V} \cup \{u\}, \quad \mathscr{E} = \mathscr{E} \cup \{(u, s^*), (u, t^*)\}$$

  **end while**
  Create a new node with index $u$ to join the 3 remaining nodes $s, t, l \in \mathscr{N}$, and update the latent tree structure $\mathscr{T}$:

$$\mathscr{V} = \mathscr{V} \cup \{u\}, \quad \mathscr{E} = \mathscr{E} \cup \{(u, s), (u, t), (u, l)\}$$

---

The next theorem shows that the neighbor joining algorithm recovers the true latent tree structure with high probability.

**Theorem 7** (Sparsistency). *Under Assumptions (A1) and (A2), let $\widehat{T}$ be the estimated tree structure using the neighbor joining algorithm and $T^*$ be the true tree structure. We define*

$$I_{\min} := \min_{s,t}\{d_{st},\ (s,t) \in T^*\}, \tag{35}$$

*where $d_{st}$ is the population tree metric. Then, under the condition that*

$$\frac{n}{k \log |\mathcal{O}|} \cdot I_{\min} \sigma_{\min}^2 \to \infty, \tag{36}$$

*we have $\liminf_{n\to\infty} \mathbb{P}\left(\widehat{T} = T^*\right) = 1$.*

From the above theorem, we see that even though our latent variable tree model is nonparametric, we get the nearly parametric scaling (we allow the dimension $|\mathcal{O}|$ to increase almost exponentially fast than the sample size $n$) for structure estimation when $k$, $I_{\min}$, and $\sigma_{\min}$ are constants.

*Proof.* Our analysis is based on the stability result of the neighbor joining algorithm. From Theorem 34 of Mihaescu et al. (2007), we have that, if the estimated tree metric $\widehat{d}_{st}$ satisfies

$$\max_{s,t} |\widehat{d}_{st} - d_{st}| \leq \frac{I_{\min}}{4}, \tag{37}$$

the neighbor joining algorithm correctly recovers the latent tree structure.

From Theorem 6, we have

$$\mathbb{P}\left(\sup_{st} \left|\widehat{d}_{st} - d_{st}\right| \leq \frac{8\kappa}{\sigma_{\min}} \cdot \sqrt{\frac{2k \log |\mathcal{O}|}{n}}\right) \geq 1 - o(1). \tag{38}$$

Therefore, it suffices if

$$\frac{8\kappa}{\sigma_{\min}} \cdot \sqrt{\frac{2k\delta \log |\mathcal{O}|}{n}} \leq \frac{I_{\min}}{4}. \tag{39}$$

This proves the desired result. □

## 9 Experiments

We evaluate our method on synthetic data as well as a real-world crime/communities dataset (Asuncion and Newman, 2007; Redmond and Baveja, 2002). For all experiments we compare to 2 existing approaches. The first is to assume the data is multivariate Gaussian and use the tree metric defined in (Choi et al., 2010) (which is essentially a function of the correlation coefficient). The second existing approach we compare to is the Nonparanormal (NPN) (Liu et al., 2009) which assumes that there exist marginal transformations $f_1, \ldots, f_p$ such that $f(X_1), \ldots, f(X_p) \sim N(\mu, \Sigma)$. If the data comes from a Nonparanormal distribution, then the transformed data are assumed to be multivariate Gaussian and the same tree metric as the Gaussian case can be used on the transformed data.

Our approach makes much fewer assumptions about the data than either of these two methods which can be more favorably in practice.

To perform inference in our approach, we use the spectral algorithm described earlier in the paper. For inference in the Gaussian (and nonparanormal) cases, we use the technique in (Choi et al., 2010) to learn the model parameters (covariance matrix). Once the covariance matrix has been estimated, marginalization in a Gaussian graphical model reduces to solving a linear equation of one variable if we are only computing the marginal of one variable given a set of evidence (Bickson, 2008).

### 9.1 Synthetic data: structure recovery.

The first experiment is to demonstrate how our method compares to the Gaussian and Nonparanormal methods in terms of structure recovery. We experiment with 3 different tree types (each with 64 leaves or observed variables): a balanced binary tree, a completely binary skewed tree (like an HMM), and randomly generated binary trees. Furthermore we explore with two types of underlying distributions: **(1)** A multivariate Gaussian with mean zero and inverse covariance matrix that respects the tree structure. **(2)** A highly non-Gaussian distribution that uses the following generative process to generate the $n$-th sample from a node $s$ in the tree (denoted $x_s^{(n)}$): If $s$ is the root, sample from a mixture of 2 Gaussians. Else, with probability $\frac{1}{2}$ sample from a Gaussian with mean $-x_{\pi_s}^{(n)}$ and with probability $\frac{1}{2}$ sample from a Gaussian with mean $x_{\pi_s}^{(n)}$.

We vary the training sample size from 200 to 100,000. Once we have computed the empirical tree distance matrix for each algorithm, we use the neighbor joining algorithm (Saitou et al., 1987) to learn the trees. For evaluation we compare the number of hops between each pair of leaves in the true tree to the estimated tree. For a pair of leaves $i, j$ the error is defined as: $error(i,j) = \frac{|hops^*(i,j) - \widehat{hops}(i,j)|}{hops^*(i,j)} + \frac{|hops^*(i,j) - \widehat{hops}(i,j)|}{\widehat{hops}(i,j)}$, where $hops^*$ is the true number of hops and $\widehat{hops}$ is the estimated number of hops. The total error is then computed by adding the error for each pair of leaves.

The performance of our method depends on the number of eigenvalues chosen and we experimented with 2, 5 and 8 singular values. Furthermore, we choose the bandwidth $\sigma$ for the Gaussian RBF kernel needed for the covariance operators using median distance between pairs of training points.

When the underlying distribution is not Gaussian, our method performs better than the Gaussian and Nonparanormal methods for all the tree structures. This is to be expected, since the non-Gaussian data we generated is neither Gaussian or Nonparamnormal, yet our method is able to learn the structure correctly. We also note that balanced binary trees are the easiest to learn while the skewed trees are the hardest (Figure 3).

Even when the underlying distribution is Gaussian, our method still performs very well compared to the Gaussian and NPN approaches and outperforms them for the binary and balanced trees. It performs worse for the skewed case likely due to the fact that the eigenvalues (dependence) decay along the length of the tree leading to larger errors in the empirical distance matrix.

### 9.2 Synthetic data: model selection.

Next we evaluate the ability of our model to select the correct number of singular values via held-out likelihood. For this experiment we use a balanced binary tree with 16 leaves (total of 31 nodes) and 100000 samples. A different generative process is used so that it is clear what the correct number of singular values should be (When the hidden state space is continuous like in our first synthetic experiment this is unclear). Each internal node is discrete and takes on $d$ values. The leaf is a mixture of $d$ gaussians where which Gaussian to sample from is dictated by the discrete value of the parent.

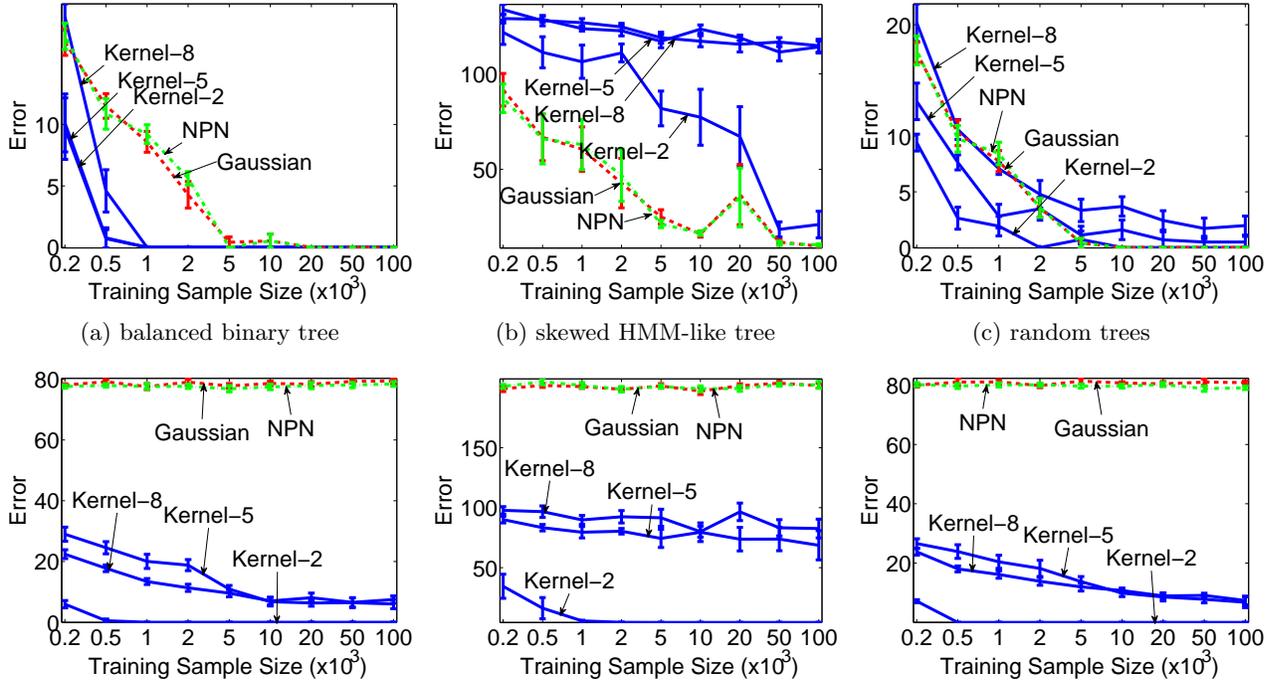

Figure 3: Comparison of our kernel structure learning method to the Gaussian and Nonparanormal methods on different tree structures. Top row: data points are generated from Gaussian distributions with latent variables connected by a tree structure. Bottom row: data points are generated from mixture of Gaussian distributions with latent variables connected by a tree structure. Especially in the latter case, our kernel structure learning method is able to adapt the data distributions and recover the structure in a much more accurate way.

We vary d from 2 through 5 and then run our method for a range of 2 through 8 singular values. We select the model that has the highest likelihood computed using our spectral algorithm on a hold-out set of 500 examples. We then take the difference between the number of singular values chosen and the true singular values, and plot histograms of this difference (Ideally all the trials should be in the zero bin). The experiment is run for 20 trials. As we can see in Figure 4, when $d$ is low, the held-out likelihood computed by our method does a fairly good job in recovering the correct number. However, as the true number of eigenvalues rises our method under-estimate the true number (although it is still fairly close).

### 9.3 Crime Dataset.

Finally, we explore the performance of our method on a communities and crime dataset from the UCI repository (Asuncion and Newman, 2007; Redmond and Baveja, 2002). In this dataset several real valued attributes are collected for several communities, such as ethnicity proportions, income, poverty rate, divorce rate etc., and the goal is to predict the number of violent crimes (proportional to size of community) that occur based on these attributes. In general these attributes are highly skewed and therefore not well characterized by a Gaussian model.

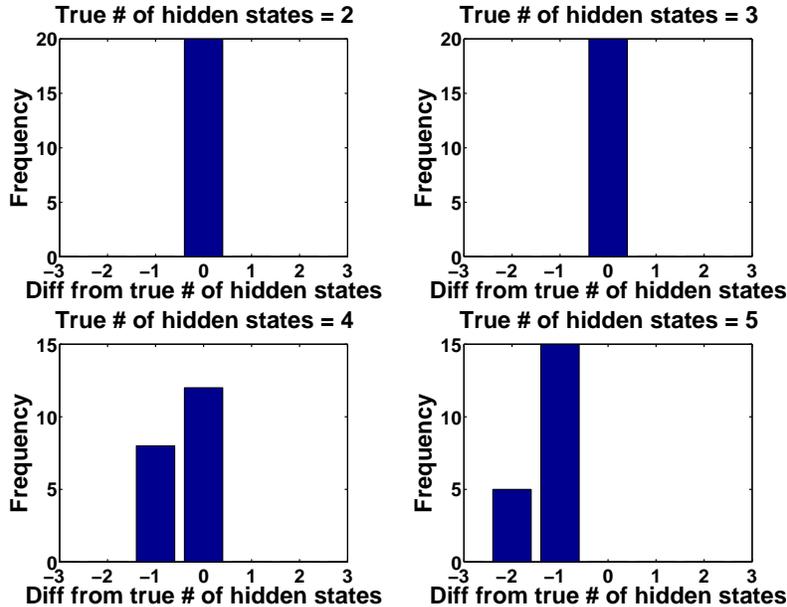

Figure 4: Histogram of the differences between the estimated number of hidden states and the true number of states.

We divide the data into 1400 samples for training, 300 samples for model selection (held-out likelihood), and 300 samples for testing. We pick the first 50 of these attributes, plus the violent crime variable and construct a latent tree using our tree metric and neighbor joining algorithm (Saitou et al., 1987). We depict the tree in Figure 5 and highlight a few coherent groupings. For example, the "elderly" group attributes are those related to retirement and social security (and thus correlated). The large clustering in the center is where the class variable (violent crimes) is located next to the poverty rate, and the divorce rate among other relevant variables. Other groupings include type of occupation and education level as well as ethnic proportions. Thus, overall our method is able to capture sensible relationships.

For a more quantitative evaluation, we condition on a set of $\mathcal{E}$ of evidence variables and predict the violent crimes class label. We experiment with a varying number of sizes of evidence sets from 5 to 40 and repeat for 40 randomly chosen evidence sets of a fixed size. Since the crime variable is a number between 0 and 1, our error measure is simply $err(\widehat{c}) = |\widehat{c} - c^*|$ (where $\widehat{c}$ is the predicted value and $c^*$ is the true value. As one can see in Figure 5 our method outperforms both the Gaussian and the nonparanormal for the range of query sizes. Thus, in this case our method is better able to capture the skewed distributions of the variables than the other methods.

## 10 Conclusions

We present a method that uses Hilbert space embeddings of distributions that can recover the latent tree structures, and perform local-mininum-free spectral learning and inference for continuous and non-Gaussian variables. Both simulation and results on real datasets show the advantage of our proposed approach for non-Gaussian data.

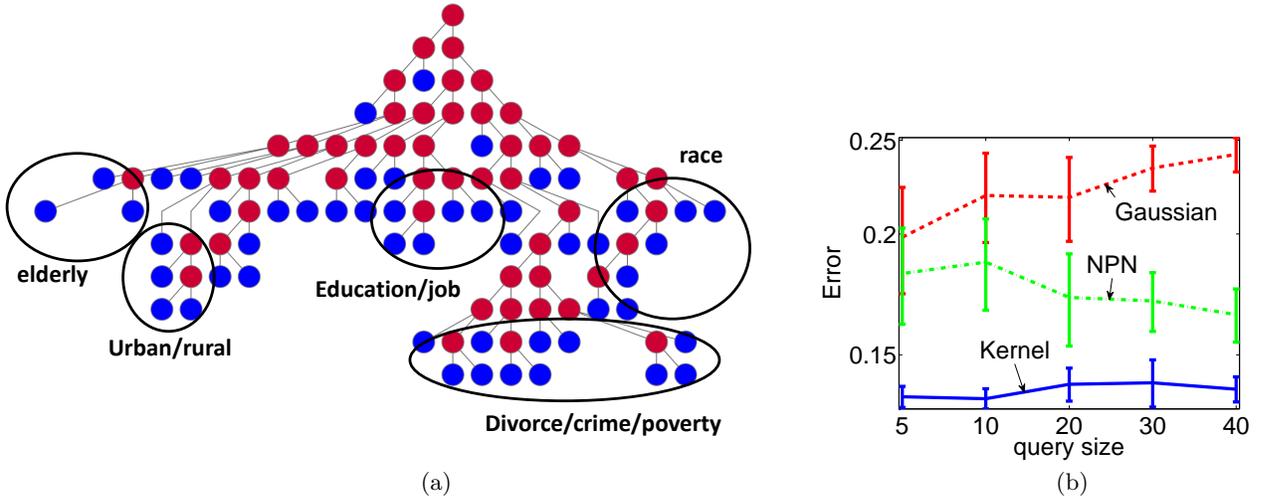

Figure 5: (a) visualization of kernel latent tree learned from crime data (b) Comparison of our method to Gaussian and NPN in predictive task.

## A  Kernel Embedding of Distributions

We begin by providing an overview of Hilbert space embedding of distributions, which are *implicit* mappings of distributions into potentially *infinite* dimensional feature spaces. A reproducing kernel Hilbert space (RKHS) $\mathcal{F}$ on $\Omega$ with a kernel $K(x, x')$ is a Hilbert space of functions $f : \Omega \mapsto \mathbb{R}$ with inner product $\langle \cdot, \cdot \rangle_{\mathcal{F}}$. Its element $K(x, \cdot)$ satisfies the reproducing property: $\langle f(\cdot), K(x, \cdot) \rangle_{\mathcal{F}} = f(x)$, and consequently, $\langle K(x, \cdot), K(x', \cdot) \rangle_{\mathcal{F}} = K(x, x')$, meaning that we can view the evaluation of a function $f$ at any point $x \in \Omega$ as an inner product. Alternatively, $K(x, \cdot)$ can be viewed as an implicit feature map $\phi(x)$ where $K(x, x') = \langle \phi(x), \phi(x') \rangle_{\mathcal{F}}$.[1] Popular kernel functions on $\mathbb{R}^d$ include the polynomial kernel $K(x, x') = (\langle x, x' \rangle + c)^r$ and the Gaussian RBF kernel $K(x, x') = \exp(-\sigma \|x - x'\|^2)$, where $\|\cdot\|$ is the Euclidean norm on the corresponding domain. Kernel functions have also been defined on graphs, time series, dynamical systems, images and other structured objects (Schölkopf et al., 2004). Thus the methodology presented below can readily be generalized to a diverse range of data types as long as kernel functions are defined for them.

### A.1  Population Definition

The Hilbert space embedding approach represents a probability distribution $P(X)$ by an element in the RKHS associated with a kernel function (Fukumizu et al., 2004; Berlinet and Thomas-Agnan, 2004; Smola et al., 2007b; Sriperumbudur et al., 2008, 2010),

$$\mu_X := \mathbb{E}_X [\phi(X)] = \int_\Omega \phi(x) \, dP(x), \tag{40}$$

---

[1] For notational simplicity, we use the same kernel for $y$, i.e., $K(y, y') = \langle \phi(y), \phi(y') \rangle_{\mathcal{F}}$

where the distribution $P(X)$ is mapped to its expected feature map, *i.e.*, to a point in a potentially infinite-dimensional and implicit feature space. The mean embedding $\mu_X$ has the property that the expectation of any RKHS function $f$ can be evaluated as an inner product in $\mathcal{F}$, i.e.,

$$\langle \mu_X, f \rangle_{\mathcal{F}} := \mathbb{E}_X[f(X)] \text{ for all } f \in \mathcal{F}.$$

Kernel embeddings can be readily generalized to the joint distributions of two or more variables using tensor product feature spaces. For instance, we can embed a joint distribution of two variables $X$ and $Y$ into a tensor product feature space $\mathcal{F} \otimes \mathcal{F}$ by

$$\mathcal{C}_{XY} := \mathbb{E}_{XY}[\phi(X) \otimes \phi(Y)] = \int_{\Omega \times \Omega} \phi(x) \otimes \phi(y) \, dP(x,y) \tag{41}$$

where we assume for simplicity that the two variables share the same domain $\Omega$ and kernel $k$, and the tensor product features satisfy $\langle \phi(x) \otimes \phi(y), \phi(x') \otimes \phi(y') \rangle_{\mathcal{F} \otimes \mathcal{F}} = K(x,x')K(y,y')$.

The joint embeddings can also be viewed as an uncentered cross-covariance operator $\mathcal{C}_{XY} : \mathcal{F} \mapsto \mathcal{F}$ by the standard equivalence between a tensor and a linear map. That is, given two functions $f, g \in \mathcal{F}$, their covariance can be computed by $\mathbb{E}_{XY}[f(X)g(Y)] = \langle f, \mathcal{C}_{XY} g \rangle_{\mathcal{F}}$, or equivalently $\langle f \otimes g, \mathcal{C}_{XY} \rangle_{\mathcal{F} \otimes \mathcal{F}}$, where in the former we view $\mathcal{C}_{XY}$ as an operator while in the latter we view it as an element in tensor product space. By analogy, $\mathcal{C}_{XX} := \mathbb{E}_X[\phi(X) \otimes \phi(X)]$ and $\mathcal{C}_{(XX)Y} := \mathbb{E}_X[\phi(X) \otimes \phi(X) \otimes \phi(Y)]$ can also be defined, $\mathcal{C}_{(XX)Y}$ can be regarded as a linear operator from $\mathcal{F}$ to $\mathcal{F} \otimes \mathcal{F}$. It will be clear from the context whether we use $\mathcal{C}_{XY}$ as an operator between two spaces or as an element from a tensor product feature space.

Kernel embeddings can be readily generalized to joint distribution of $O$ variables, $X_1, \ldots, X_O$, using the $O$-th order tensor product feature space $\mathcal{F}^O$ (Here $\mathscr{O} := \{X_1, \ldots, X_O\}$). In this feature space, the feature map is defined as $\otimes_{i=1}^O \phi(x_i) := \phi(x_1) \otimes \phi(x_2) \otimes \ldots \otimes \phi(x_O)$, and the inner product in this space satisfies $\langle \otimes_{i=1}^O \phi(x_i), \otimes_{i=1}^O \phi(x'_i) \rangle_{\mathcal{F}^O} = \prod_{i=1}^O \langle \phi(x_i), \phi(x'_i) \rangle_{\mathcal{F}} = \prod_{i=1}^O K(x_i, x'_i)$. Then we can embed a joint density $p(x_1, \ldots, x_O)$ into a tensor product feature space $\mathcal{F}^O$ by

$$\mathcal{C}_{\mathscr{O}} := \mathbb{E}_{\mathscr{O}}\left[\otimes_{i=1}^O \phi(X_i)\right] = \int_{\Omega^O} \left(\otimes_{i=1}^O \phi(x_i)\right) \, dP(x_1, \ldots, x_O). \tag{42}$$

Although the definition of embedding in (40) is simple, it turns out to have both rich representational power and a well-behaved empirical estimate. First, the mapping is injective for characteristic kernels (Sriperumbudur et al., 2008, 2010). That is, if two distributions, $P(X)$ and $Q(Y)$, are different, they will be mapped to two distinct points in the feature space. Many commonly used kernels are characteristic, such as the Gaussian RBF kernel $\exp(-\sigma\|x-x'\|^2)$ and Laplace kernel $\exp(-\sigma\|x-x'\|)$, which implies that if we embed distributions using these kernels, the distance of the mappings in feature space will give us an indication on whether these two distributions are identical. This intuition has been exploited to design state-of-the-art two-sample tests (Gretton et al., 2012) and independence tests (Gretton et al., 2008). For the former case, the test statistic is the squared distance between the embeddings of $P(Y)$ and $Q(Y)$, i.e., $\texttt{mmd}(X,Y) := \|\mu_X - \mu_Y\|_{\mathcal{F}}^2$. For the latter case, the test statistic is the squared distance between the embeddings of a joint distribution $P(X,Y)$ and the product of its marginals $P(X)P(Y)$, i.e., $\texttt{hsic}(X,Y) := \|\mathcal{C}_{XY} - \mu_X \otimes \mu_Y\|_{\mathcal{F} \otimes \mathcal{F}}^2$. Similarly, this statistic also has advantages over the Kernel density etiolation based statistic. We will further discuss these tests in the next section, following our introduction of finite sample estimates of the distribution embeddings and test statistics.

## A.2 Finite Sample Kernel Estimator

While we rarely have access to the true underlying distribution, $P(X)$, we can readily estimate its embedding using a finite sample average. Given a sample $\mathcal{D}_X = \{x_1, \ldots, x_n\}$ of size $n$ drawn i.i.d. from $P(X)$, the empirical Hilbert space embedding is

$$\widehat{\mu}_X = \frac{1}{n} \sum_{i=1}^n \phi(x_i). \tag{43}$$

This empirical estimate converges to its population counterpart in RKHS norm, $\|\widehat{\mu}_X - \mu_X\|_{\mathcal{F}}$, with a rate of $O_p(n^{-\frac{1}{2}})$ (Berlinet and Thomas-Agnan, 2004; Smola et al., 2007b). We note that this rate is independent of the dimension of $X$, meaning that statistics based on Hilbert space embeddings circumvent the curse of dimensionality.

Kernel embeddings of joint distributions inherit the previous two properties of general embeddings: injectivity and easy empirical estimation. Given $m$ pairs of training examples $\mathcal{D}_{XY} = \{(x_1, y_1), \ldots, (x_n, y_n)\}$ drawn i.i.d. from $P(X, Y)$, the covariance operator $\mathcal{C}_{XY}$ can then be estimated as

$$\widehat{\mathcal{C}}_{XY} = \frac{1}{n} \sum_{i=1}^n \phi(x_i) \otimes \phi(y_i). \tag{44}$$

By virtue of the kernel trick, most of the computation required for statistical inference using Hilbert space embeddings can be reduced to the Gram matrix manipulation. The entries in the Gram matrix $K$ correspond to the kernel value between data points $x_i$ and $x_j$, i.e., $K_{ij} = K(x_i, x_j)$, and therefore its size is determined by the number of data points in the sample (similarly Gram matrix $G$ has entries $G_{ij} = K(y_i, y_j)$). The size of the Gram matrices is in general much smaller than the dimension of the feature spaces (which can be infinite). This enables efficient nonparametric methods using the Hilbert space embedding representation. For instance, the empirical `mmd` can be computed using kernel evaluations,

$$\widehat{\text{mmd}}(P, Q) = \left\| \frac{1}{n} \sum_{i=1}^n \phi(x_i) - \frac{1}{n} \sum_{i=1}^n \phi(y_i) \right\|_{\mathcal{F}}^2 = \frac{1}{n^2} \sum_{i,j=1}^n \left( K(x_i, x_j) + K(y_i, y_j) - 2K(x_i, y_j) \right).$$

If the sample size is large, the computation in Hilbert space embedding methods may be expensive. In this case, a popular solution is to use a low-rank approximation of the Gram matrix, such as incomplete Cholesky factorization Fine and Scheinberg (2001), which is known to work very effectively in reducing computational cost of kernel methods, while maintaining the approximation accuracy.

# B Kernel Embeddings of Conditional Distributions

In this section, we define the Hilbert space embeddings of conditional distributions, which take into account complex conditional independence relations in graphical models.

## B.1 Population Definition

The Hilbert space embedding of a conditional distribution $P(Y|X)$ is defined as (Song et al., 2009)

$$\mu_{Y|x} := \mathbb{E}_{Y|x}[\phi(Y)] = \int_\Omega \phi(y) \; dP(y|x). \tag{45}$$

Given this embedding, the conditional expectation of a function $g \in \mathcal{F}$ can be computed as
$$\mathbb{E}_{Y|x}[g(Y)] = \langle g, \mu_{Y|x} \rangle_{\mathcal{F}}.$$
This may be compared with the property of the mean embedding in the previous section, where the *unconditional* expectation of a function may be written as an inner product with the embedding. Unlike the embeddings discussed in the previous section, an embedding of conditional distribution is not a single element in the RKHS, but will instead sweep out a family of points in the RKHS, each indexed by a fixed value $x$ of the conditioning variable $X$. It is only by fixing $X$ to a particular value $x$, that we will be able to obtain a single RKHS element, $\mu_{Y|x} \in \mathcal{F}$. In other words, we need to define an operator, denoted as $\mathcal{C}_{Y|X}$, which can take as input an $x$ and output an embedding. More specifically, we require it to satisfy
$$\mu_{Y|x} = \mathcal{C}_{Y|X}\phi(x). \qquad (46)$$

Based on the relation between conditional expectation and covariance operators, Song et al. (2009) show that that, under the assumption $\mathbb{E}_{Y|\cdot}[g(Y)] \in \mathcal{F}$,
$$\mathcal{C}_{Y|X} := \mathcal{C}_{YX}\mathcal{C}_{XX}^{-1}, \quad \text{and hence } \mu_{Y|x} = \mathcal{C}_{YX}\mathcal{C}_{XX}^{-1}\phi(x) \qquad (47)$$
satisfy the requirement in (46). We remark that the assumption $\mathbb{E}_{Y|\cdot}[g(Y)] \in \mathcal{F}$ always holds for finite domains with characteristic kernels, but it is not necessarily true for continuous domains (Fukumizu et al., 2004). In the cases where the assumption does not hold, we will use the expression $\mathcal{C}_{YX}\mathcal{C}_{XX}^{-1}\phi(x)$ as an approximation of the conditional mean $\mu_{Y|x}$. In practice, the inversion of the operator can be replaced by the regularized inverse $(\mathcal{C}_{XX} + \mathcal{I})^{-1}$, where $\mathcal{I}$ is the identity operator.

## B.2 Finite Sample Kernel Estimator

Given a dataset $\mathcal{D}_{XY} = \{(x_1, y_1), \ldots, (x_n, y_n)\}$ of size $n$ drawn i.i.d. from $P(X, Y)$, we will estimate the conditional embedding operator as
$$\widehat{\mathcal{C}}_{Y|X} = \Phi(\mathbf{K} + \lambda \mathbf{I})^{-1}\Upsilon^\top \qquad (48)$$
where $\Phi := (\phi(y_1), \ldots, \phi(y_n))$ and $\Upsilon := (\phi(x_1), \ldots, \phi(x_n))$ are implicitly formed feature matrix, $\mathbf{K} = \Upsilon^\top \Upsilon$ is the Gram matrix for samples from variable $X$, and $\mathbf{I}$ is an identity matrix. Furthermore, we need an additional regularization parameter $\lambda$ to avoid overfitting. Then $\widehat{\mu}_{Y|x} = \widehat{\mathcal{C}}_{Y|X}\phi(x)$ becomes a weighted sum of the feature mapped data points $\phi(y_1), \ldots, \phi(y_n)$,
$$\widehat{\mu}_{Y|x} = \sum_{i=1}^{m} \beta_i(x)\phi(y_i) = \Phi\boldsymbol{\beta}(x) \quad \text{where} \qquad (49)$$
$$\boldsymbol{\beta}(x) = (\beta_1(x), \ldots, \beta_n(x))^\top = (\mathbf{K} + \lambda \mathbf{I})^{-1}\mathbf{K}_{:x},$$
and $\mathbf{K}_{:x} = (K(x, X_1), \ldots, K(x, X_n))^\top$. The empirical estimator of the conditional embedding is similar to the estimator of the ordinary embedding from equation (43). The difference is that, instead of applying uniform weights $\frac{1}{n}$, the former applies *non-uniform* weights, $\beta_i(x)$, on observations which are, in turn, determined by the value $x$ of the conditioning variable. These non-uniform weights reflect the effects of conditioning on the embeddings. It is also shown that this empirical estimate converges to its population counterpart in RKHS norm, $\|\widehat{\mu}_{Y|x} - \mu_{Y|x}\|_{\mathcal{F}}$, with rate of $O_p(n^{-\frac{1}{4}})$ if one decreases the regularization $\lambda$ with rate $O(n^{-\frac{1}{2}})$. With appropriate assumptions on the joint distribution of $X$ and $Y$, better rates can be obtained (Grunewalder et al., 2012).

# C  Hilbert Space Embeddings as Infinite Dimensional Higher Order Tensors

The above Hilbert space embedding $\mathcal{C}_{\mathscr{O}}$ can also be viewed as a multi-linear operator (tensor) of order $O$ mapping from $\mathcal{F}^O$ to $\mathbb{R}$. (More detailed and generic introduction to tensor and tensor notation can be found in Kolda and Bader (2009)). The operator is linear in each argument (mode) when fixing other arguments. Furthermore, the application of the operator to a set of elements $\{f_i \in \mathcal{F}\}_{i=1}^O$ can be defined using the inner product from the tensor product feature space, i.e.,

$$\mathcal{C}_{\mathscr{O}} \bullet_1 f_1 \bullet_2 \ldots \bullet_O f_O := \langle \mathcal{C}_{\mathscr{O}}, \otimes_{i=1}^O f_O \rangle_{\mathcal{F}^O} = \mathbb{E}_{\mathscr{O}} \left[ \prod_{i=1}^O \langle \phi(X_i), f_i \rangle_{\mathcal{F}} \right], \tag{50}$$

where $\bullet_i$ means applying $f_i$ to the $i$-th argument of $\mathcal{C}_{\mathscr{O}}$. Furthermore, we can define Hilbert-Schmidt norm $\|\cdot\|_{HS}$ of $\mathcal{C}_{\mathscr{O}}$ as $\|\mathcal{C}_{\mathscr{O}}\|_{HS}^2 = \sum_{i_1=1}^{\infty} \cdots \sum_{i_O=1}^{\infty} (\mathcal{C}_{\mathscr{O}} \bullet_1 e_{i_1} \bullet_2 \ldots \bullet_O e_{i_O})^2$ using an orthonormal basis $\{e_i\}_{i=1}^{\infty} \subset \mathcal{F}$ [2]. The Hilbert-Schmidt norm can be viewed as a generalization of the Frobenius norm from matrices to general operators. For the space of operators with finite Hilbert-Schmidt norms, we can also define the Hilbert-Schmidt inner product of such operators as

$$\left\langle \mathcal{C}_{\mathscr{O}}, \widetilde{\mathcal{C}}_{\mathscr{O}} \right\rangle_{HS} = \sum_{i_1=1}^{\infty} \cdots \sum_{i_O=1}^{\infty} (\mathcal{C}_{\mathscr{O}} \bullet_1 e_{i_1} \bullet_2 \ldots \bullet_O e_{i_O})(\widetilde{\mathcal{C}}_{\mathscr{O}} \bullet_1 e_{i_1} \bullet_2 \ldots \bullet_O e_{i_O}). \tag{51}$$

When $\mathcal{C}_{\mathscr{O}}$ has the form of $\mathbb{E}_{\mathscr{O}} \left[ \otimes_{i=1}^O \phi(X_i) \right]$, the above inner product reduces to $\mathbb{E}_{\mathscr{O}}[\widetilde{\mathcal{C}}_{\mathscr{O}} \bullet_1 \phi(X_1) \bullet_2 \ldots \bullet_O \phi(X_O)]$.

In this paper, the ordering of the tensor modes is not essential. Therefore we simply label them using the corresponding random variables. We can reshape a higher order tensor into a lower order one by partitioning its modes into several disjoint groups. For instance, let $\mathscr{I}_1 = \{X_1, \ldots, X_s\}$ be the set of modes corresponding to the first $s$ variables and $\mathscr{I}_2 = \{X_{s+1}, \ldots, X_O\}$. Similarly to the Matlab notation, we can obtain a 2nd order tensor by

$$\mathcal{C}_{\mathscr{I}_1;\mathscr{I}_2} = \text{reshape}\left(\mathcal{C}_{\mathscr{O}}, \mathscr{I}_1, \mathscr{I}_2\right) \,:\, \mathcal{F}^s \times \mathcal{F}^{O-s} \mapsto \mathbb{R}. \tag{52}$$

To under the reshape operator, if we have a set of functions $\{f_i\}_{i=1}^O$, then

$$\mathcal{C}_{\mathscr{O}} \bullet_1 f_1 \bullet_2 f_2 \bullet_3 \ldots \bullet_O f_O = \mathcal{C}_{\mathscr{I}_1;\mathscr{I}_2} \bullet_1 (f_1 \otimes f_2 \otimes \ldots \otimes f_s) \bullet_2 (f_{s+1} \otimes f_{s+2} \otimes \ldots \otimes f_O).$$

Note that given an orthonormal basis $\{e_i\}_{i=1}^{\infty} \in \mathcal{F}$, we can readily obtain an orthonormal basis for $\mathcal{F}^s$ as $\{e_{i_1} \otimes \ldots \otimes e_{i_s}\}_{i_1,\ldots,i_s=1}^{\infty}$, and for $\mathcal{F}^{O-s}$ as $\{e_{i_1} \otimes \ldots \otimes e_{i_{O-s}}\}_{i_1,\ldots,i_{O-s}=1}^{\infty}$. Hence we can define the Hilbert-Schmidt norm for $\mathcal{C}_{\mathscr{I}_1;\mathscr{I}_2}$. Using (50), it is easy to see that $\|\mathcal{C}_{\mathscr{I}_1;\mathscr{I}_2}\|_{HS} = \|\mathcal{C}_{\mathscr{O}}\|_{HS}$.

In the reverse direction, we can also reshape a lower order tensor into a higher order one by further partitioning certain mode of the tensor. For instance, we can partition $\mathscr{I}_1$ into $\mathscr{I}_1' = \{X_1, \ldots, X_t\}$ and $\mathscr{I}_1'' = \{X_{t+1}, \ldots, X_s\}$, and turn $\mathcal{C}_{\mathscr{I}_1;\mathscr{I}_2}$ into a 3rd order tensor by

$$\mathcal{C}_{\mathscr{I}_1';\mathscr{I}_1'';\mathscr{I}_2} = \text{reshape}\left(\mathcal{C}_{\mathscr{I}_1;\mathscr{I}_2}, \mathscr{I}_1', \mathscr{I}_1'', \mathscr{I}_2\right) \,:\, \mathcal{F}^t \times \mathcal{F}^{s-t} \times \mathcal{F}^{O-s} \mapsto \mathbb{R}, \tag{53}$$

---

[2] The definition of Hilbert-Schmidt norm is invariant to arbitrary orthonormal basis.

meaning that if we have a set of functions $\{f_i\}_{i=1}^{O}$, then

$$\mathcal{C}_{\mathscr{I}_1;\mathscr{I}_2} \bullet_1 (f_1 \otimes f_2 \otimes ... \otimes f_s) \bullet_2 (f_{s+1} \otimes f_{s+2} \otimes ... \otimes f_O)$$
$$= \mathcal{C}_{\mathscr{I}_1';\mathscr{I}_1'';\mathscr{I}_2} \bullet_1 (f_1 \otimes f_2 \otimes ... \otimes f_t) \bullet_2 (f_{t+1} \otimes f_2 \otimes ... \otimes f_s) \bullet_3 (f_{s+1} \otimes f_{s+2} \otimes ... \otimes f_O),$$

and furthermore, we have $\left\| \mathcal{C}_{\mathscr{I}_1';\mathscr{I}_1'';\mathscr{I}_2} \right\|_{HS} = \| \mathcal{C}_{\mathscr{O}} \|_{HS}$.

The 2nd order tensor $\mathcal{C}_{\mathscr{I}_1;\mathscr{I}_2}$ are the same as the cross-covariance operator between two sets of variables in $\mathscr{I}_1$ and $\mathscr{I}_2$. In this case, we adopt the same notation and operations as for matrices. For instance, we can perform singular value decomposition of $\mathcal{C}_{\mathscr{I}_1;\mathscr{I}_2} = \sum_{i=1}^{\infty} s_i (u_i \otimes v_i)$ where $s_i \in \mathbb{R}$ are ordered in nonincreasing manner, $\{u_i\}_{i=1}^{\infty} \subset \mathcal{F}^s$ and $\{v_i\}_{i=1}^{\infty} \subset \mathcal{F}^{d-s}$ are left and right singular vectors. The rank of $\mathcal{C}_{\mathscr{I}_1;\mathscr{I}_2}$ is the smallest $r$ such that $s_i = 0$ for $i \geq r$. In this case, we will also define $\mathcal{U}_r = (u_1, u_2, \ldots, u_r)$, $\mathcal{V}_r = (v_1, v_2, \ldots, v_r)$ and $\mathcal{S}_r = \mathrm{diag}\,(s_1, s_2, \ldots, s_r)$, and denote the low rank representation as $\mathcal{C}_{\mathscr{I}_1;\mathscr{I}_2} = \mathcal{U}_r \mathcal{S}_r \mathcal{V}_r^\top$. We denote this to be "thin" SVD. Finally, $\mathrm{reshape}\,(\mathcal{C}_{\mathscr{O}}, \{\mathscr{O}\}, \emptyset)$, a 1st order tensor, is simply a vector where we will use vector notation.

# References

ANANDKUMAR, A., CHAUDHURI, K., HSU, D., KAKADE, S. M., SONG, L. and ZHANG, T. (2011). Spectral methods for learning multivariate latent tree structure. *arXiv preprint arXiv:1107.1283* .

ANANDKUMAR, A. and VALLUVAN, R. (2013). Learning loopy graphical models with latent variables: Efficient methods and guarantees. *The Annals of Statistics* **41** 401–435.

ASUNCION, A. and NEWMAN, D. (2007). Uci machine learning repository.

BERLINET, A. and THOMAS-AGNAN, C. (2004). *Reproducing Kernel Hilbert Spaces in Probability and Statistics*. Kluwer.

BICKSON, D. (2008). Gaussian belief propagation: Theory and application. *Arxiv preprint arXiv:0811.2518* .

BLEI, D., NG, A. and JORDAN, M. (2002). Latent dirichlet allocation. In *Advances in Neural Information Processing Systems 14* (T. G. Dietterich, S. Becker and Z. Ghahramani, eds.). MIT Press, Cambridge, MA.

CHOI, M., TAN, V., ANANDKUMAR, A. and WILLSKY, A. (2010). Learning latent tree graphical models. *Arxiv preprint arXiv:1009.2722* .

CHOW, C. and LIU, C. (1968). Approximating discrete probability distributions with dependence trees. *IEEE Transactions on Information Theory* **14** 462–467.

CLARK, A. (1990). Inference of haplotypes from PCR-amplified samples of diploid populations. *Molecular Biology and Evolution* **7** 111–122.

DASKALAKIS, C., MOSSEL, E. and ROCH, S. (2006). Optimal phylogenetic reconstruction. In *Proceedings of the thirty-eighth annual ACM symposium on Theory of computing*. ACM.

Daskalakis, C., Mossel, E. and Roch, S. (2011). Phylogenies without branch bounds: Contracting the short, pruning the deep. *SIAM Journal on Discrete Mathematics* **25** 872–893.

Dempster, A. P., Laird, N. M. and Rubin, D. B. (1977). Maximum likelihood from incomplete data via the EM algorithm. *Journal of the Royal Statistical Society B* **39** 1–22.

Erdös, P. L., Székely, L. A., Steel, M. A. and Warnow., T. J. (1999a). A few logs suffice to build (almost) all trees (I). *Random Structures and Algorithms* **14** 153–184.

Erdös, P. L., Székely, L. A., Steel, M. A. and Warnow., T. J. (1999b). A few logs suffice to build (almost) all trees: Part II. *Theoretical Computer Science* **221** 77–118.

Fine, S. and Scheinberg, K. (2001). Efficient SVM training using low-rank kernel representations. *Journal of Machine Learning Research* **2** 243–264.

Fukumizu, K., Bach, F. R. and Jordan, M. I. (2004). Dimensionality reduction for supervised learning with reproducing kernel Hilbert spaces. *Journal of Machine Learning Research* **5** 73–99.

Gretton, A., Borgwardt, K., Rasch, M., Schoelkopf, B. and Smola, A. (2012). A kernel two-sample test. *Journal of Machine Learning Research* **13** 723–773.

Gretton, A., Fukumizu, K., Teo, C.-H., Song, L., Schölkopf, B. and Smola, A. J. (2008). A kernel statistical test of independence. In *Advances in Neural Information Processing Systems 20*. MIT Press, Cambridge, MA.

Gronau, I., Moran, S. and Snir, S. (2008). Fast and reliable reconstruction of phylogenetic trees with very short edges. In *Proceedings of the nineteenth annual ACM-SIAM symposium on Discrete algorithms*. Society for Industrial and Applied Mathematics.

Grunewalder, S., Lever, G., Baldassarre, L., Patterson, S., Gretton, A. and Pontil, M. (2012). Conditional mean embeddings as regressors. In *Proceedings of the International Conference on Machine Learning*.

Harmeling, S. and Williams, C. (2010). Greedy learning of binary latent trees. *IEEE Transactions on Pattern Analysis and Machine Intelligence* .

Heller, K. and Ghahramani, Z. (2005). Bayesian hierarchical clustering. In *Proceedings of the 22nd international conference on Machine learning*. ACM.

Hoff, P. D., Raftery, A. E. and Handcock, M. S. (2002). Latent space approaches to social network analysis. *Journal of the American Statistical Association* **97** 1090–1098.

Kolda, T. G. and Bader, B. W. (2009). Tensor decompositions and applications. *SIAM Review* **51** 455–500.

Liu, H., Lafferty, J. and Wasserman, L. (2009). The nonparanormal: Semiparametric estimation of high dimensional undirected graphs. *The Journal of Machine Learning Research* **10** 2295–2328.

Mossel, E. (2007). Distorted metrics on trees and phylogenetic forests. *IEEE/ACM Transactions on Computational Biology and Bioinformatics (TCBB)* **4** 108–116.

Mossel, E. and Roch, S. (2006). Learning nonsingular phylogenies and hidden markov models. *Annals of Applied Probability* **16** 583–614.

Mossel, E., Roch, S. and Sly, A. (2011a). On the inference of large phylogenies with long branches: How long is too long? *Bulletin of mathematical biology* **73** 1627–1644.

Mossel, E., Roch, S. and Sly, A. (2011b). Robust estimation of latent tree graphical models: Inferring hidden states with inexact parameters. *IEEE Transactions on Information Theory* .

Parikh, A., Song, L. and Xing, E. P. (2011). A spectral algorithm for latent tree graphical models. In *Proceedings of the International Conference on Machine Learning*.

Parzen, E. (1962). On estimation of a probability density function and mode. *The Annals of Mathematical Statistics* **33** 1065–1076.

Pearl, J. (2001). *Causality: Models, Reasoning and Inference*. Cambridge University Press.

Rabiner, L. R. and Juang, B. H. (1986). An introduction to hidden Markov models. *IEEE ASSP Magazine* **3** 4–16.

Redmond, M. and Baveja, A. (2002). A data-driven software tool for enabling cooperative information sharing among police departments. *European Journal of Operational Research* **141** 660–678.

Roch, S. (2010). Toward extracting all phylogenetic information from matrices of evolutionary distances. *Science* **327** 1376–1379.

Rosenblatt, M. (1956). Remarks on some nonparametric estimates of a density function. *The Annals of Mathematical Statistics* 832–837.

Saitou, N., Nei, M. et al. (1987). The neighbor-joining method: a new method for reconstructing phylogenetic trees. *Mol Biol Evol* **4** 406–425.

Schölkopf, B., Tsuda, K. and Vert, J.-P. (2004). *Kernel Methods in Computational Biology*. MIT Press, Cambridge, MA.

Semple, C. and Steel, M. (2003). *Phylogenetics*, vol. 24. Oxford University Press, USA.

Shawe-Taylor, J. and Cristianini, N. (2004). *Kernel Methods for Pattern Analysis*. Cambridge University Press, Cambridge, UK.

Silverman, B. W. (1986). *Density Estimation for Statistical and Data Analysis*. Monographs on statistics and applied probability, Chapman and Hall, London.

Smola, A., Gretton, A., Song, L. and Schölkopf, B. (2007a). A hilbert space embedding for distributions. In *Algorithmic Learning Theory* (E. Takimoto, ed.). Lecture Notes on Computer Science, Springer.

Smola, A. J., Gretton, A., Song, L. and Schölkopf, B. (2007b). A Hilbert space embedding for distributions. In *Proceedings of the International Conference on Algorithmic Learning Theory*, vol. 4754. Springer.

Song, L., Gretton, A. and Guestrin, C. (2010). Nonparametric tree graphical models. In *Proc. Intl. Conference on Artificial Intelligence and Statistics.*

Song, L., Huang, J., Smola, A. J. and Fukumizu, K. (2009). Hilbert space embedding of conditional distributions. In *Proceedings of the International Conference on Machine Learning.*

Sriperumbudur, B., Gretton, A., Fukumizu, K., Lanckriet, G. and Schölkopf, B. (2008). Injective Hilbert space embeddings of probability measures. In *Proc. Annual Conf. Computational Learning Theory.*

Sriperumbudur, B., Gretton, A., Fukumizu, K., Lanckriet, G. and Schölkopf, B. (2010). Hilbert space embeddings and metrics on probability measures. *Journal of Machine Learning Research* **11** 1517–1561.

Wasserman, L. (2006). *All of Nonparametric Statistics.* Springer.

Zhang, N. (2004). Hierarchical latent class models for cluster analysis. *The Journal of Machine Learning Research* **5** 697–723.